# Machine Learning for E-mail Spam Filtering: Review, Techniques and Trends

**Alexy Bhowmick · Shyamanta M. Hazarika**

**Abstract** We present a comprehensive review of the most effective content-based e-mail spam filtering techniques. We focus primarily on Machine Learning-based spam filters and their variants, and report on a broad review ranging from surveying the relevant ideas, efforts, effectiveness, and the current progress. The initial exposition of the background examines the basics of e-mail spam filtering, the evolving nature of spam, spammers playing cat-and-mouse with e-mail service providers (ESPs), and the Machine Learning front in fighting spam. We conclude by measuring the impact of Machine Learning-based filters and explore the promising offshoots of latest developments.

## 1 Introduction

Electronic-mail (abbreviated as *e-mail*) is a fast, effective and inexpensive method of exchanging messages over the Internet. Whether its a personal message from a family member, a company-wide message from the boss, researchers across continents sharing recent findings, or astronauts staying in touch with their family (via e-mail uplinks or IP phones), e-mail is a preferred means for communication. Used worldwide by 2.3 billion users, at the time of writing the article, e-mail usage is projected to increase up to 4.3 billion accounts by the year-end 2016 [Radicati, 2016]. But the increasing dependence on e-mail has induced the emergence of many problems caused by 'illegitimate' e-mails, i.e. *spam*. According to the Text Retrieval Conference (TREC) the term '*spam*' is - an unsolicited, unwanted e-mail that was sent indiscriminately [Cormack, 2008]. Spam e-mails are unsolicited, un-ratified and usually mass mailed. Spam being a carrier of malware causes the proliferation of unsolicited advertisements, fraud schemes, phishing messages, explicit content, promotions of cause, etc. On an organizational front, spam effects include: *i)* annoyance to individual users, *ii)* less reliable e-mails, *iii)* loss of work productivity, *iv)* misuse of network bandwidth, *v)* wastage of file server storage space and computational power, *vi)* spread of viruses, worms, and Trojan horses, and *vii)* financial losses through phishing, Denial of Service (DoS), directory harvesting attacks, etc.[Siponen and Stucke, 2006].

Over the couple of decades e-mail spam volume has increased exponentially and is not just an annoyance but a security threat; as it continues to evolve in its potential to do serious damage to individuals, businesses and economies. The fact that e-mail is a very cheap means of reaching to millions of potential customers serves as a strong motivation for amateur advertisers and direct marketers [Cranor and Lamacchia, 1998]. For e.g. one of the favorite spam topics is the '*penny stock*' spam or the *pump and dump* schemes that take place over the Internet platform. Fraudsters (spammers) purchase large quantities of '*penny stocks*' i.e. stocks of small, thinly traded companies, through compromised brokerage accounts and promote them via message boards or abroad e-mail campaign, pointing to the transient increase in share value. Even if a fraction of the recipients are fooled into buying the stocks, the spammers make a huge profit. Unwitting investors seeking higher gains believe the hype and purchase the

Alexy Bhowmick · Shyamanta M. Hazarika
School of Engineering,
Tezpur University
Tezpur, Assam, India.
E-mail: alexyb@tezu.ernet.in

stocks, creating higher demand and raising the price further. Soon after the spam e-mails are sent, the fraudster sells off his stocks at premium leaving the duped investors desperate to sell their own. Stock spam is just an old trick that has made a massive comeback in the first half of 2013. The Security Threat Report 2014 [Sophos, 2014] suggests that on some days, 50% of the overall spam volume were '*pump and dump*' mailings.

According to recent reports [IBM, 2014], [Cyberoam, 2014], [Symantec, 2014], spam is being increasingly used to distribute viruses, malware, links to phishing sites, etc. An average of 54 billion spam e-mails was sent worldwide each day [Cyberoam, 2014]. Sizeable chunks were that of pharmacy spam, dating spam, online product purchase, diet products and online casinos spam [CISCO, 2014]. Another kind of spam that is rapidly evolving is '*Political spam*'; e-mail, contrary to popular media such as print, radio, or television provides political contestants an economical medium to get through to broad constituents of the electorate. Political spam is but a campaign tactic that mostly involves marketing for political ends or mudslinging.

Spam is a broad concept that is still not completely understood. In general, spam has many forms - chat rooms are subject to *chat spam*, blogs are subject to *blog spam* (splogs)[Kolari et al, 2006], search engines are often misled by *web spam* (search engine spamming or spamdexing)[Gyongyi and Garcia-molina, 2005], [Shi and Xie, 2013], while social systems are plagued by *social spam* [Lee et al, 2010a]. This paper focuses on '*e-mail spam*' and its variants, and not 'spam' in general. Prior attempts to review e-mail spam filtering using Machine Learning have been made, the most notable ones being [Androutsopoulos et al, 2006], [Carpinter and Hunt, 2006], [Blanzieri and Bryl, 2008], [Cormack, 2008], [Guzella and Caminhas, 2009], and Wang et al [2013]. We extend earlier surveys by taking an updated set of works into account. We consider e-mail header analysis and analysis of non-content features, which were not discussed in the fairly recent overviews by [Guzella and Caminhas, 2009] and [Wang et al, 2013], who have performed topic modeling instead. We also present a content analysis of the major spam-filtering surveys over the period (2004-2015). Significant amounts of historical and recent literature, including gray literature (dissertations, press articles, technical and security reports, web publications, etc.) were studied to report recent advances and findings. We believe our survey is of complementary nature and provides an inclusive survey of the state-of-the-art in content-based e-mail spam filtering.

Our work addresses the following:

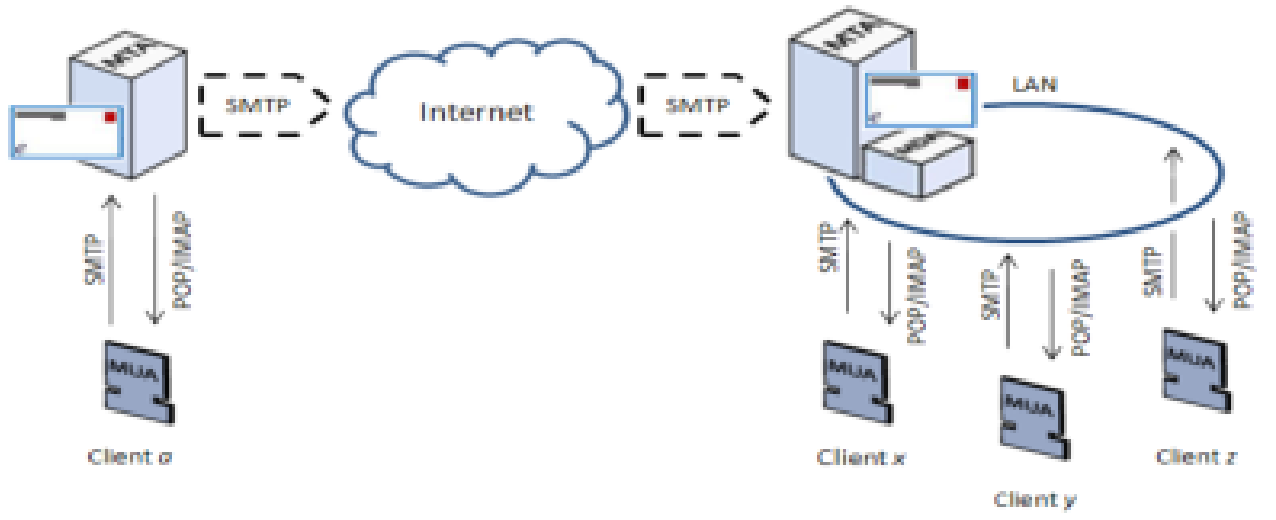

**Fig. 1** The e-mail architecture.

- *First*, we perform an extensive evolutionary exploration of the major spam characteristics, trends and spammers evasion techniques. In doing so, we underline some promising research directions and a few research gaps.
- *Second*, we discuss feature engineering for textual and image spam e-mails. We investigate alternate spam filtering plans based on e-mail header and non-content features.
- *Third*, we present taxonomy of content-based e-mail spam filtering and a qualitative summary of major surveys on spam e-mails over the period (2004-2015).
- *Fourth*, we report new findings and suggest lines of future investigations into machine learning techniques for emerging spam types.

The rest of the paper is organized as follows: In Sec 2, we characterize spam evolution, trends, spam causes and their counter measures. In Sec 3, we discuss corpus pre-processing, feature extraction, feature selection and analysis of header and non-content features. In Sec 4, we review spam filtering techniques employed prior to Machine Learning. Section 5 offers details on Machine Learning algorithms applied successfully to textual and multimedia content of spam e-mails. Special attention is given to recent techniques. Section 6 overviews standard evaluation measures and publicly available e-mail spam, image spam and phishing e-mail corpuses. Finally, Section 7 outlines future research trends.

### 1.1 E-mail and Spam Filters

When an e-mail is sent, it enters into the messaging system and is routed from one server to another till it reaches the recipients mailbox. Figure 1 depicts the e-mail architecture and how e-mail works. E-mail depends on few primary protocols: SMTP (Simple Mail Transfer Protocol) [Jonathan B. Postel, 1982], POP3 (Post Office Protocol) and IMAP (Internet Message Access Protocol). The transmission details are specified by the SMTP protocol. POP3 and IMAP are the most widely implemented protocols for the Mail User Agent (MUA) and are basically used to receive messages. A Message Transfer Agent (MTA) receives mails from a sender MUA or some other MTA and then determines the appropriate route for the mail [Katakis et al, 2007]. The recipients MTA delivers the incoming mail to the incoming mail server Mail Delivery Agent (MDA) which is basically a POP/IMAP server. MUAs (e.g. Mozilla Thunderbird, Microsoft Outlook, etc.) are *e-mail clients* and help the user to read and write e-mails. Spam filters can be deployed at strategic places in both clients and servers. Many Internet Service Providers (ISPs) and organizations deploy spam filters at the e-mail server level, the preferred places to deploy being at the gateways, mail routers, etc. They can be deployed in clients, where they can be installed at proxies or as plug-ins, as in [Irwin and Friedman, 2008]. Some spam filters, (e.g. *SpamBayes*) can be deployed at both server and client levels.

### 1.2 Structure of an E-mail

An e-mail comprises of two elements: *body* and the *header*. The e-mail body comprises of unstructured data such as text, HTML markup, multimedia objects and attachments. The header comprises trace information and structured fields that are part of the message content. The Simple Mail Transfer Protocol (SMTP) [Jonathan B. Postel, 1982] defines e-mail header session to contain fields like - the subject, senders name, e-mail ID, sending date, routing information, timestamp, etc. for recipient information and successful delivery. Each attribute (*field*) in the header has a name and specific meaning -

- **Received:** *Contains transit-related information of e-mail servers, IP addresses, dates, etc.*
- **From:** *Sender's name; e-mail ID.* "Name" <e-mail@example.com>
- **To:** *Recipient's name; e-mail ID.* "Name" <e-mail@example.com>
- **Return Path:** *Encloses an optional address specification to be used if an error is encountered (bounce).*
- ***Message ID:*** *A single unique message identifier designated by the mail system.*
- **X-mailer:** *The mail software used to create/send the message.*
- **Subject:** *String identifies the theme of the message placed by the sender.*
- **Content type:** *Format of content (character set, etc.), specified by MIME (Multipurpose Internet Mail Extensions).*

Each e-mail message comprises of the transit-handling *envelope* [Crocker, 2009] that is hidden from e-mail users. First the *envelope* sender address is sent, followed by one or more envelope recipient addresses, and finally the actual message is sent. The e-mail servers actually use the envelope address (not the message header address) to deliver the e-mail to the correct recipient. The final recipient sees only the e-mail header and body. The envelope address is one of the e-mail features that is very often abused by spammers.

## 2 Characterizing Spam Evolution

A couple of decades earlier spam e-mail content was mainly textual. Therefore, spam filters analyzed only the e-mail body and header to distinguish *ham* (legitimate e-mails) from *spam* e-mails. Today however, amateur advertisers and opportunists harness addresses from chat rooms, web pages, newsgroup archives, service provider directories etc and send junk e-mail blindly to millions without much cost [Androutsopoulos et al, 2006]. Anti-spam software companies and research groups working on spam filtering for quite some time now have tasted limited success, mostly because spam filtering is an *adversarial classification* task. In such tasks, a malicious adversary '*poisons*' the training data with carefully crafted attack techniques in order to mislead a classifier [Jorgensen et al, 2008]. To deliver spam e-mail to a huge number of recipients, spammers often resort to use of bulk mailing software or e-mail harvesters [Blanzieri and Bryl, 2008].

Spam evolution has been briefly discussed in scientific literature [Carpinter and Hunt, 2006] [Guzella and Caminhas, 2009] [Almeida and Yamakami, 2012]. One reason why spam is difficult to filter is because of its dynamic nature. The characteristics (*e.g.* topics, frequent terms, etc) of spam e-mail vary rapidly over time as spammers always seek to invent new strategies to bypass spam filters. These strategies include - word obfuscation, image spam, sending e-mail spam from hijacked computers, etc. A proper understanding of the spam nature and evolution can help much in the development of proper countermeasures. Some of the evasion techniques and major trends in spam causes and characteristics seen over the years are discussed below:

### 2.1 Word Obfuscation

Words like '*sex*', '*free*', '*congratulations*' are good indicators of spam and have large ('spammy') weights. Initial spam filters based on heuristic filtering could easily detect and filter spam e-mails based on the presence of such obvious words. Figure 2 illustrates a word cloud of common words in spam e-mail [Greenberg, 2010]; the

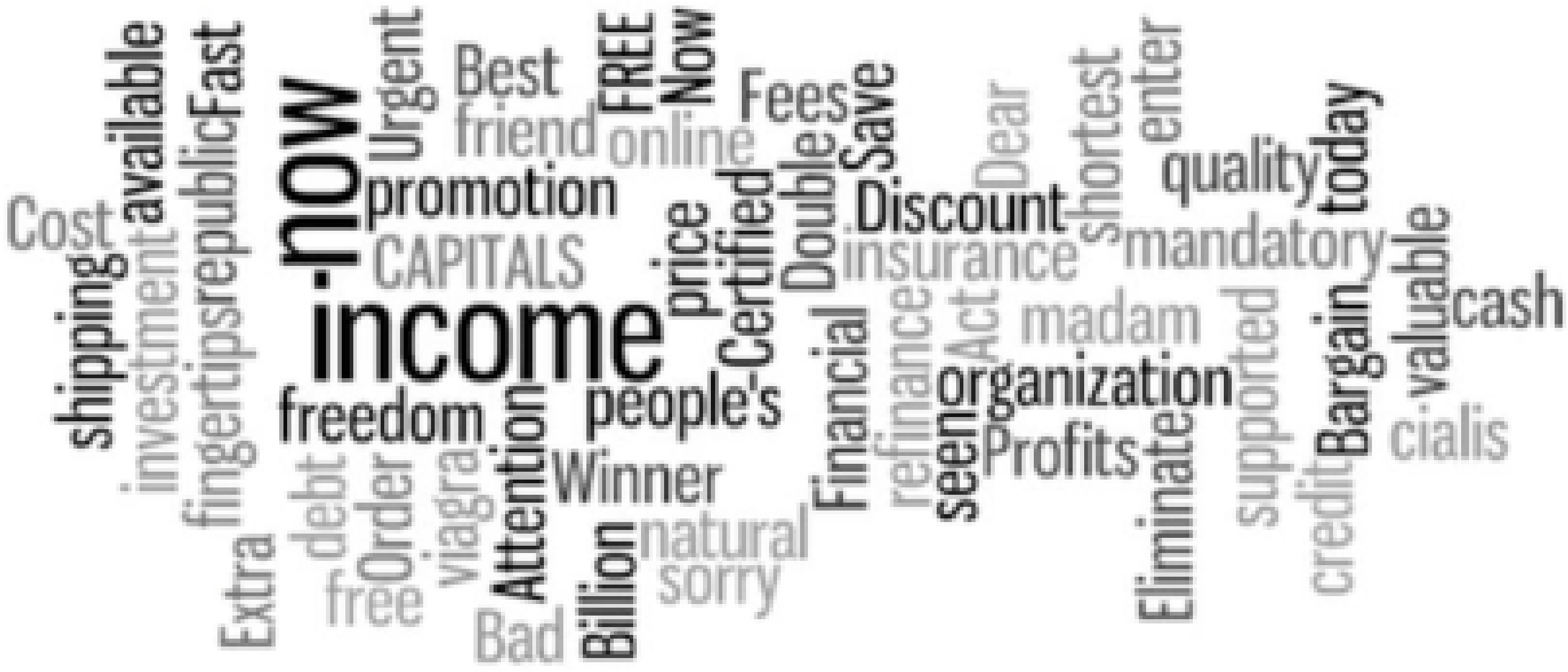

**Fig. 2** A word cloud of common words in spam e-mail.

larger a word appears, the more often it has been found to occur in e-mail spam). Spammers adapted quickly by making sure such obvious words are not encountered verbatim in their messages. To defeat filters, they resorted to simple obfuscation techniques like breaking the word into multiple pieces, as -

- `f-r-e-e`
  embedding special characters
- `fr<!--xx-->ee`
  using HTML comments
- `\item <a href='mailto:%66ree'>free</a>`
  with character-entity encoding
- `\item o fr&#101xe`
  encoded with HTML ASCII codes

When seen by any web user, all these above samples look the same as "*free*" but they thwart simple word/phrase filtering and escape the filter rules. The effectiveness of filter re-training however caused spammers to abandon one technique and migrate to newer obfuscation techniques. HTML-based obfuscation techniques are discussed at length in the study by [Pu and Webb, 2006]. Spammers resorted to the use of innocuous words to obfuscate the e-mail message content in order to confuse or circumvent spam filters. In general, there are many ways to obscure the e-mail content: misplaced spaces, purposeful misspellings, embedded special characters (letter substitution), Unicode letter transliteration [Liu and Stamm, 2007], HTML redrawing, etc. Tokenization attacks are a similar spamming technique more associated with the preprocessing stage in spam filtering. In tokenization the spammer works to defeat the feature selection process by splitting and modifying the crucial message features. Examples include introducing spaces, special symbols, asterisks in words or HTML, JavaScript, CSS layout tricks. A classic example of evading the recognition of the word 'VIAGRA by the spam filter would be 'V-I-A-G-R-A'.

## 2.2 Bayesian Poisoning Attacks

A usual criticism of statistical spam filters (e.g. *SpamBayes*,*DSPAM*, *POPFile*) is that they are susceptible to '*poisoning*' by interjection of random words into the spam messages [Fawcett, 2004], [Graham-Cumming, 2006]. *Bayesian poisoning* is such a kind of statistical attack in which spammers use carefully crafted e-mails to attack the heart of a Bayesian filter and thus degrade its effectiveness. The spammers add random or carefully selected legitimate-seeming words in order to confuse the spam filter and cause it to believe an incoming spam e-mail is not spam (a *statistical II error*). Spammers can get these common English words or Ham phrases from sources like - Reuters news articles, written and spoken English, and USENET messages. These strong statistical attacks have an incidental consequence too - *a statistical I error* or simply a higher false positive rate. The reason is that when the user trains the spam filter with the poisoned training data, the spam filter 'learns' about such random words as being good evidences of spam [Sanz, 2008]. Paul Graham [Graham, 2002b] however played down the effectiveness of such poisoning techniques arguing that to outweigh the statistical significance of even one incriminating word as "viagra", spammers would need many innocent words (e.g. names of ones friends and family, terms used at work, etc) which are unique for each recipient and spammers have no way of figuring them out. However, evidence suggests Bayesian poisoning is real and cannot be dismissed [Biggio et al, 2011].

Graham-Cumming [Graham-Cumming, 2004] [Graham-Cumming, 2006], identified two types of possible attacks on Bayesian filters: *passive* (where in absence of feedback, the spammer can at best make educated guesses) and *active* (where the spammer discovers an effective wordlist after getting feedback). [Lowd and Meek, 2005] investigated '*good word attacks*' where a spammer appends words indicative of legitimate e-mail, and found Naive Bayes extremely vulnerable to both scenarios of active and passive attacks. Their results showed frequent filter re-training could mitigate the effective of these attacks. [Wittel and Wu, 2004] explored a simple passive attack of poisoning with random words (a *dictionary attack*) and found it ineffective against CRM114[1], but effective against SpamBayes.[2] A smarter passive attack with common or 'hammy' words (*common word* or *focused attack*) saw SpamBayes perform even worse but CRM114 remained very resistant. [Stern et al, 2004] showed that injecting common words from the English language led to the performance decrease of SpamBayes. Published research indicates that Bayesian poisoning is real and the number of published attack methods indicates that it cannot be dismissed and that further investigation on poisoning of statistical spam filters is a worthwhile task of research.

### 2.3 Backscatter Spam

When an e-mail is sent, the sender is normally informed if the e-mail could not be delivered or the delivery was delayed for some reason. E-mail servers normally send a bounce message notifying the sender of delivery problems. Such a message is termed - *Delivery Status Notification* (DSN). Mostly, DSNs are welcome to the sender and they are generally sent to the envelope sender address. Backscatter occurs when DSNs are sent to senders whose addresses are forged in the message envelope by spammers. In other words, backscatter are delivery notifications from another server, rejecting an e-mail made to come across as being mailed from an account [Cormack and Lynam, 2007]. These mails are then delivered unsolicited in bulk quantities to a lot of recipients. Hence, backscatter qualifies as unsolicited bulk e-mail and is spam. Misdirected bounces from mail servers, misdirected "*please confirm your subscription*" requests from mailing lists, "*out of office*" vacation autoreplies and auto-responders, challenge requests from Challenge/Request Systems, etc. are the major varieties of backscatter. Backscatter, also called '*collateral spam*' is a direct consequence of spam.

[Cormack and Lynam, 2007] experimented with six open-source filters and a test set of 49,086 messages with backscatter representing a mere 1% of the total spam in the test set. It was found that content-based spam filters could filter 98% of the spam, but backscatter was found to be most difficult to classify with nearly all the backscatter messages being misclassified. Backscatter is a problem that is hard to deal with and though spammers may be blamed for it, it simply exists because our mail servers are configured to bounce messages back to fake addresses rather than just reject such spam immediately [McMillan, 2008]. Servers that generate e-mail backscatter can land up on various DNS-based Blacklists (DNSBLs). Improperly configured e-mail servers gives rise to '*open relays*' which contribute to the problem of backscatter. Open relay servers can also get listed in various DNSBLs.

### 2.4 Image Spam

Text-based spam filters are designed only to analyze different components of an e-mail (sender's address, header, body, attachments) and detect specific spam characteristics. A new type of spam called image-based spam or *image spam* is a rapidly spreading. It involves textual spam content embedded into images that are attached to e-mails. OCR (Optical Character Recognition)-based modules are effective to a limited extent against image-spam [Biggio et al, 2006] [Fumera, 2006]. But often the textual content is obfuscated by spammers to evade OCR tools. Till 2010, the upsurge of spam e-mails meant that roughly up to 85% of all e-mail spam were image spam [Wu and Tsai, 2008].

SpamAssassin, a widely used commercial and open-source spam filter provides several OCR plug-ins (e.g. OCR Plugin [3], Fuzzy OCR Plugin[4], and Bayes OCR Plugin[5]) that can be used to detect image spam. It has been established from current literature that the applying modern classification approaches to the generated text from image spam is very efficient. Later, signatures were also generated to easily detect and filter already known image spam. The spam filter database by mid-2012 contained more than 40 million relevant spam signatures [IBM, 2012]. In order to avoid signature-based techniques, spammers switched tactics by making arbitrary alterations to a specified template image. They began employing obfuscation, similar to the approach usually applied in web forums, to outsmart Optical Character Recognition (OCR) tools. Lately, Pattern Recognition techniques and Computer Vision are playing a significant part in filtering of multimedia data. However, the solutions achieved so far have shortcomings, and their efficiency is yet to be systematically investigated [Biggio et al, 2006].

[1] CRM 114 - the Controllable Regex Mutilator, an open-source spam filtering device

[2] SpamBayes - a popular open-source spam filtering tool, with 700,000 downloads, is based on techniques laid out by Paul Graham.

[3] A SpamAssassin OCR plug-in is maintained at: http://wiki.apache.org/spamassassin/OcrPlugin

[4] Fuzzy OCR is available but no longer maintained.

[5] A beta version of BayesOCR plugin is available at http://pralab.diee.unica.it/en/BayesOCR

Image-based filtering involves extraction of relevant features from the image and classification by state-of-the-art classifiers. Image-based spam detection is an example of classification of multimedia data. A number of researchers have devised approaches based on Pattern Recognition and Computer Vision to address different forms of image spam. In general they can be grouped into *two* broad categories: *a)* OCR-based techniques and *b)* Low level image features based techniques. The use of OCR tools to extract text embedded into images, and processing it using modern text categorization techniques was thoroughly investigated by [Fumera, 2006]. But OCRs have been proven to be computationally expensive and not accurate enough in adversarial situations [Goodman et al, 2007] [Attar et al, 2011]. [Biggio et al, 2006] surveyed and categorized the major techniques which have been suggested as image spam solutions. Spilling of image spam onto social networks like Twitter or Facebook has become widespread. Extraction of features for image-based spam filters is further discussed in Sec 3. A detailed and recent review involving definitions, spam tricks, complete classification of image spam filtering techniques and datasets may be found in [Wu and Tsai, 2008].

## 2.5 Botnet Spam

At a time when blacklists had almost put the spammers out of business and diminished their profits, some enterprising spammers joined hands with virus and exploit code writers to get access to compromised machines on the Internet known as '*bots*' or '*zombies*'. The term *botnet* applies to an army of machines that are compromised and controlled by a single '*botmaster*'. A bot, when subverted (*e.g.* by a virus/Trojan infection or by a specific bot software), can be used to send out spam or malware, harvest password and login information for identity theft and fraud, re-route users to spoofed websites, or even recruit new bots, and so on. Botnets on the other hand constitute a major threat to the Internet infrastructure as they have the capability to - mount crippling denial of service (DoS) attacks on servers, generate click-fraud [Perera et al, 2013], send out a flood of spam and backscatter [Xie et al, 2008] facilitate phishing and *pump-and-dump* schemes, form a computational grid to break weak passwords or obfuscate the operators point of origin, etc. Botnets run on the global level outside the range of national boundaries. According to public tracker Shadowserver [6], at least one million zombie machines or *bots* are believed to be active and the number is still growing.

Identifying and blacklisting each and every bot is challenging, both because a botnet attack is momentary and the fact that a single bot transmits only a small volume of spam e-mails to avoid detection. On the other hand spammers are using large Botnets to send spam, thus creating extremely a huge number of IP addresses to be blacklisted. *Grum*, a sneaky, kernel-mode rootkit was of notable interest to researchers. It was a relatively small botnet with only 600,000 members. Yet it was responsible for almost 25 percent, or 40 billion spam e-mails a day before it was finally taken down. Identifying botnets is a new challenge for the anti-spam industry, and tracking spammers and bringing them to justice, and pulling down botnet servers becomes an international undertaking. July 18, 2012 saw the take down of the Grum botnet [Sophos, 2013]. Recently, as a repercussion of the bombing incident during the Boston Marathon which happened on April 15, 2013, botnet spam related to the Boston Marathon bombing was found to have constituted 40 percent of all spam messages transmitted globally on subsequent days [CISCO, 2014].

According to CISCO report [CISCO, 2007], botnets are the primary security threat on the Internet today. Botnets are hard to detect because of their dynamic nature and their adaptability in evading the common security defenses. Botnets have been studied thoroughly, particularly in the context of spam and phishing [Xie et al, 2008], [John et al, 2009] and [Zhuang et al, 2008]. Botnets are emerging as the most severe threat against cyber-security as they provide a distributed platform for several unlawful activities like distributed denial of service (DDoS) attacks, malware dissemination, phishing, scanning and click fraud. Because botnets attack from multiple fronts there is no single technology that can provide protection from it.

[6] https://www.shadowserver.org/wiki/

### 2.6 Social Engineering - *Phishing*

Spammers are increasingly adopting the use of social engineering techniques in the spam campaign. Patient and committed attackers perform extensive research and gain a sophisticated understanding of the needs and motivation of recipients and then contact them with highly believable communications (*e.g.* e-mails or social networking message) which may reflect knowledge of the individuals' work activities, colleagues, friends, and family. *Phishin*g is an illegal attempt that exploits both social engineering and technical deception to acquire sensitive confidential data (*e.g.* social security number, e-mail address, passwords, etc.) and financial account credentials [Robinson, 2003] [Bergholz et al, 2010]. Phishing involves spam e-mails disguised as legitimate with a subject or message designed to trick the victims into revealing confidential information. In *deceptive phishing*, e-mail notifications appearing to come from credit card companies, security agencies, banks, providers, online payment processors or IT administrators are commonly used to exploit the unsuspecting public. The notification encourages the recipient to urgently enter/update their personal data. In most cases, the fraudsters try to frighten a recipient by some "urgent" matter (e.g. "We suspect an unauthorized transaction on your account. To ensure that your account is not compromised, please click the link below and confirm your identity") that requires their immediate attention and divulging of their personal information. It is often accompanied by a threat to block the account within a limited period, if not responded. Once information such as user-name and password are entered, it becomes a clear case of identity theft followed by worse consequences such as transfer of cash from a victims account, official documents being obtained, or goods being purchased using stolen credentials. Malicious users are also interested in other types of passwords, such as those for social networks, e-mail accounts and other services [Kaspersky, 2014]. In *malware-based phishing*, malicious software is spread through e-mails or by exploiting security loopholes and installed on the user's machine. The malware may then capture user inputs, and confidential information may be sent to the 'phisher' [Bergholz et al, 2010]. The phishers' top targets in 2012 were social networks, financial institutions, non-profit organizations and search engines [IBM, 2014] [Kaspersky, 2014].

Phishing attacks use e-mail as their main carrier in order to allure unmindful victims. Phishing can also occur on a fake web site that is a perfect replica of the official site, such as the log-in page for a banking web site, to harvest e-mail addresses and log-on credentials of their victims. The companies spoofed most often were found to be Barclays, Bank of America, PayPal, eBay etc [Ludl et al, 2007]. Phishing attacks and identity theft-based scams are becoming more sophisticated in their exploitation of social engineering techniques. While spamming affects bandwidth; social engineering attacks like phishing directly affect their victims. In recent years '*pharming*' has evolved to be a major concern to e-commerce and banks sites. In '*Pharming*' the attacker redirects unsuspecting users to fake sites or proxy servers with seeded scripts [Abu-nimeh et al, 2007] [Kaspersky, 2014]. The Internet Security Threat Report [Symantec, 2014] states that in 2013 the rate of phishing had increased, from 1 in 414 for 2012 to 1 in 392 in 2013. Much of these phishing attempts involve the creation of fake login pages for popular social networks sites. Besides spoofing login pages of legitimate sites, phishers also began launching baits relevant to current events for flavouring the phishing pages.

Several browser extensions (*e.g.* SafeCache and SafeHistory for Mozilla) and plug-ins (e.g. SpoofGuard) have been proposed [Chou et al, 2004], [Stepp, 2005], [Nattakant, 2009] and [Sta, 2014]. [Chandrasekaran et al, 2006] have pointed out several weaknesses of existing browser-based solutions and proposed a novel Support Vector Machine (SVM) - based technique for e-mail spam filtering based on the inherent structural properties in phishing e-mails. [Abu-nimeh et al, 2007] evaluated the predictive accuracy of six popular machine learning-based classifiers on phishing data sets. Phishing countermeasures such as secure e-mail authentication, password hashing, etc. involves high administrative overhead, hence *content-based filtering* can be used to detect phishing attacks and improve existing solutions. While we agree client-side solutions for phishing have been developed over the years even by huge software companies, server-side solutions are the focus of research [Abu-nimeh et al, 2007], [Fette et al, 2007] and [Basnet et al, 2008].

Bergholz et al [2010] have identified a number of highly informative features about phishing attempts and also proposed a server-side statistical phishing filter. The success of phishing is largely determined by the low levels of user-awareness regarding how the fraudsters and spoof sites operate. Increasing user awareness will help them to learn to spot the telltale signs of social engineering tricks, which includes, undue pressure, a false sense of urgency, bogus official letters, too-good-to-be-true offers, quid-pro-quo offers, etc. Meanwhile spam filters remain the first line of defense against phishing. According to Anti-Phishing Working Group [Anti-Phishing Working Group (APWG), 2014] new brands continue to be targeted by phishers and to battle these phishing attacks, presently the world needs more phishing databases.

## 3 Corpus Preprocessing

Not all information present in an e-mail is necessary or useful. Eliminating the less informative and noisy terms lowers the feature space dimensionality and enhances classification performance in most cases [Guzella and Caminhas, 2009], [Diao et al, 2003] and [Shi et al, 2012]. Corpus preprocessing is a process that involves transforming the mail corpus into a uniform format that is more comprehensible to the machine learning algorithms [Zhang et al, 2004], [Katakis et al, 2007]. Due to the adversarial nature of spam, spam filters need to constantly adapt to changing spam tactics, particularly in feature extraction and feature selection aspects. No matter which learning strategy is chosen for the training and testing of content-based filters, it is extremely crucial to handcraft a private corpus or use a corpus that is publicly available. In any case, e-mails need to undergo preprocessing as a preparation for feature extraction. Furthermore, a corpus may have an immense number of features, it is very important to choose features judiciously so as to prevent the classifiers from over-fitting [Drucker et al, 1999]. The effectiveness and success of content-based spam filters depends on - *feature engineering* i.e. defining and creating those features more likely to make the classifier perform better. The primary steps involved in extraction of features from an e-mail are -

- ***Lexical Analysis (Tokenization)***: The string of text representing a message is tokenized in order to identify the candidate words to be adopted as relevant spam or ham terms. Headers, attachments, and HTML tags are stripped, leaving behind just the e-mail body and subject line text. IP addresses and domain names can also be considered as *tokens*.
- ***Stop-word Removal***: Stop-word removal involves removing frequently used non-informative words, e.g. '*a*', '*an*', '*the*', and '*is*', etc. Obscure texts or symbols may also be removed in subsequent steps. Stop-word removal makes the selection of candidate terms more efficient and reduces the feature space considerably.
- ***Stemming***: Word-stemming is a term used to describe a process of converting words to their morphological base forms, mainly eliminating plurals, tenses, gerund forms, prefixes and suffixes. Stemming is closely related to lemmatization which while reducing a word considers the part of speech and the context of the word. The primary advantages of employing word stemming and lemmatization are feature space dimension reduction and classifier accuracy.
- ***Representation***: Involves the conversion of an e-mail message into a specific or structured format as needed by the machine learning algorithm being employed.

[Androutsopoulos et al, 2000a] studied the effect of corpus size, lemmatization, and stop-lists while in [Androutsopoulos et al, 2000c], they studied the effect of word stemming and stop-word removal on the performancce of classifiers. Their results show that often they do not contribute to much improvement over the filters without them. [Chih-Chin Lai and Tsai, 2004] found that stemming did not introduce any significant improvement in the filter's performance, though it did reduce the feature set size. On the contrary, employing stopping produced better performance.

### 3.1 Extracting Features

The easiest feature extraction method is the *bag of words* (BOW) model (or *vector-space model*), in which words occurring in the e-mail are treated as features. Given a set of terms T $= \{t_1, t_2, t_3 ... t_n\}$, the *bag of words* model represents a document $d$ as an N-dimensional feature vector $x = \{x_1, x_2, x_3 ... x_n\}$ where $x_i$ is a function of the occurrence of $t_i$ in $d$. It is possible to use all the features for classification. However a feature selection mechanism may be applied to select the best N features by some measure and thus reduce dimensionality. Another simple text representation is the *bag of character n-grams*. [Kanaris et al, 2006] investigated on character *n-grams* and words in spam filtering to demonstrate the advantage of *n-grams* over word-tokens. Sparse Binary Polynomial Hashing (SBPH) [Yerazunis, 2003] is another feature generator from e-mails. However, its many features made it computationally heavy and of limited use. Siefkes et al [2004] proposed an effective feature combination technique known as the Orthogonal Sparse Bigrams (OSB) to extract more compact features. Experiments showed that OSB slightly performed better than SBPH with regard to error rate. Recently, [Zhu and Tan, 2011] proposed a feature extraction approach based on local concentration (LC) which efficiently extracted position-correlated information from e-mail messages. For each style of e-mail analysis, a spam filter developer must decide on a way for performing feature extraction.

## 3.2 Feature Selection

Feature selection is a key issue and has become the subject of much research. It has a three-fold objective: *i)* enhancing the prediction accuracy of the classifiers, *ii)* building faster and economical classifiers, and *iii)* obtaining a better understanding of the elementary process involved in generation of data [Guyon, 2003]. Dimensionality reduction and feature subset selection are two preferred techniques for lowering the feature set dimension. While feature subset selection involves the extraction of a subset of the original attributes, dimensionality reduction involves linear combinations of the original feature set [Gansterer and Ecker, 2008]. [Cormack, 2008] suggests *stop-word removal* as a trivial example of feature selection, and *stemming* as a simple example of dimensionality reduction. Information Gain (IG) is one of the simplest and most successful techniques for feature selection. As discussed earlier, natural language processing provides different feature selection ways, the simplest being the '*bag of words*' model coupled with '*stemming*' and '*stopping*'. [Zhang et al, 2004] investigated the impact of three popular feature selection techniques - Document Frequency (DF), Information Gain (IG) and $\chi^2$ (CHI) test. A novel feature selection method named Comprehensively Measure Feature Selection (CMFS) was presented and evaluated with popular feature selection methods - Information Gain (IG), Chi statistic (CHI), Document Frequency (DF), Orthogonal Centroid Feature Selection (OCFS) and DIA association factor (DIA) to demonstrate that the new method notably outperformed them all [Yang et al, 2012]. Several well-known methods for feature selection are explained and compared with new feature selection methods in [Yang and Pedersen, 1997], [Yang et al, 2011] and [Gomez et al, 2012]. [Toolan and Carthy, 2010] address the issue of effective feature selection by exploring the utility of over 40 features (extracted from ham, spam and phishing pages) that have been used in recent literature.

## 3.3 E-mail Header Analysis

E-mail headers determine the recipient of a message and record the specific route the message takes as it passes through each mail server. Message headers are very reliable and powerful sources containing discriminative features for spam filtering besides the Subject and e-mail content. In fact, experimental results confirm that the e-mail header provides powerful cues for machine learning algorithms to efficiently filter out spam e-mails [Chih-Chin Lai and Tsai, 2004], [Zhang et al, 2004], [Sheu, 2007] and [Wang and Chen, 2007]. This fact was unknown in spam filtering research before and much research focused on the e-mail message body only.

**Fig. 3** The header of a typical e-mail.

According to [Zhang et al, 2004], a spam filter trained using header features alone can achieve better or comparable results than the body solution. Statistical analysis by [Wang and Chen, 2007] showed that 92.5% of 10,024 junk e-mails were filtered out using the header features - message-ID, mail user agent (MUA), sender address etc. [Hu et al, 2010] and [Al-jarrah et al, 2012] note performance evaluation of several header-based spam classifiers and evaluated their performance in filtering e-mail spam. [Sheu, 2007] mined association rules out of other basic attributes in the e-mail header sessions and proposed an efficient decision tree-based spam filtering method. E-mail Header analysis has evolved to be a very promising research area. As a filter technique it has the capability to provide low false positive rates either by itself or when used with other anti-spamming techniques.

## 3.4 Filters Based on Non-content Features

Much research has been accomplished in e-mail classification proposing general and specific solutions to the spam problem. However, most of these approaches explored only the content-based features [Drucker et al, 1999], [Androutsopoulos et al, 2000b], [Androutsopoulos et al, 2000a] and [Sakkis et al, 2001]. Filtering based solely on e-mail content has been argued to be a fundamentally flawed idea. Although such content-based methods have been effective, the perfectly malleable content of an e-mail and spammers reactivity to filtering methods gives rise to many challenges [mentioned in Section 2.1 and 2.2]. Different features such as temporal information, message length, MIME content type, proportion of symbols in e-mail body, presence of attachments, number of URLs in the e-mail, etc., are considered non-content features, and have led to promising results in differentiating incoming e-mails. Non-content features may include header features such as '*originator field*', '*destination field*', '*X-mailer field*' etc. but they are not limited to header features. [Hu et al, 2010], [Hershkop and Stolfo, 2005], and [Wang et al, 2005] describe exploiting non-content features for profiling e-mails and developing efficient and scalable non-content based spam-filtering frameworks. Table 1 illustrates the popular approaches for feature extraction and feature selection adopted by researchers and their key inferences.

**Table 1** A summary of Feature Extraction and Feature Selection techniques in popular literature.

| Authors | Approaches |
|---|---|
| [Zhang et al, 2004] | Studied subject line, header, and message body.<br>Employed Information Gain (IG), Document Frequency (DF), and $\chi^2$ test (CHI) for feature selection.<br>Found *bag of words* model quite effective on spam filtering, and header features as important as message body. |
| [Kanaris et al, 2006] | Extracted character *n-grams* of fixed length, Variable-length character n-grams<br>Explored Information Gain (IG) as a feature selection technique<br>Character *n-grams* were noted to be richer and definitve than word-tokens. |
| [Delany and Bridge, 2006] | Considered features of three types: *word, character, structured features.* in a feature-based *vs* feature-free comparison.<br>Employed Information Gain (IG) as a feature selection technique<br>Noted feature-free methods to be more correct than the feature-based system, however feature-free approaches took much longer than feature-based approach in classifying e-mails. |
| [Yeh et al, 2005] | Used behavioral patterns of spammers, Metaheuristics as features<br>Employed Term Frequency, Inverse Document Frequency (TFIDF), SpamKANN for feature selection<br>Tested SVM, Decision trees, Naive Bayes to get increased prediction accuracy than keywords. |
| [Diao et al, 2003] | Experimented on features: Header (H), Textual (T), handcrafted features (HH), etc.<br>Different ways of feature selection for Decision Tree and Naive Bayes models were evaluated<br>The usefulness and importance of different type of features were discussed in detail in experiments. |
| [Méndez et al, 2006] | Considered subject, body, header, attachment feature.<br>Analyzed strength and weaknesses of Document frequency (DF), Information Gain (IG) and $\chi^2$ test (CHI), Mutual Information.<br>Presented a deep analysis of feature selection methods. Found e-mail attachments to be useful when integrated with models. |

#### *3.4.1 Analyzing Temporal Features*

As a novel solution to the spam problem, [Kiritchenko et al, 2004] employed temporal features of an e-mail to the conventional content-based approaches to create a richer information space to work with. A simple example of a temporal feature, obtainable from the message header timestamps, is the day of the week or the time of the day the e-mail was received. They represented temporal information in the form of temporal patterns, presented an algorithm for mining temporal patterns in an e-mail sequence and described approaches to integrate temporal patterns into content-based e-mail classification. [Hao et al, 2009] explored various spatio-temporal features of e-mail senders and investigated ways to deduce the reputation of an e-mail sender based only on such features. To improve the state of affairs they presented SNARE - a sender reputation system with robust classification accuracy. Investigations reveal that the use of temporal features to improve spam filter accuracy is perhaps one of the most uncharted territories in spam filtering research.

#### *3.4.2 SMTP Path Analysis*

SMTP path analysis operates by learning about the '*spamminess*' or goodness of IP addresses by examining the history of e-mail delivered through that IP address. SMTP traffic analysis when used in combination with traditional filters does improve the accuracy of the filters. [Leiba et al, 2005] established that examining IP addresses was useful and presented a new algorithm for learning the reputation of e-mail domains and IP addresses by examining the SMTP path used to transmit e-mails. Beverly and Sollins [Beverly and Sollins, 2008] examined a variety of SMTP flow characteristics and developed a spam classifier '*SpamFlow*' based on the statistical discriminatory power of these flows.

#### *3.4.3 Behavior Analysis*

The behavioral pattern of an e-mail is 'what the sender does in composing or distributing e-mails'. Legitimate e-mails have mostly normal and meaningful behavioral patterns, while spam e-mails have abnormal or even conflicting behavior patterns. [Yeh et al, 2005] considered behavior patterns such as data spoofing, time anomaly, relay anomaly, etc; and described them by meta-heuristics and employed them as features for the classification task. To recognize spam and viruses as irregular behviour in the e-mail, [Hershkop, 2006] proposed some behavior models, some of them are recipient frequency, group communication, user's past activity histogram, etc. [Ramachandran and Feamster, 2006] studied the spammer's behaviour at the network level and found that most spam was received from a small number of regions of IP address space. They suggested that filtering based on network-level characteristics would be much more effective to combat spam as network-level properties are less malleable than e-mail content. [Li et al, 2007] performed an experimental study of the community behavior of spammers and came up with various clustering structures among their population. Based on those structures they proposed some group-based anti-spam strategies exploiting group membership of perceived spam sources. Further work on investigating clustering structures of spammers based on features as - Content length, Time of arrival, Frequency of e-mail, etc. was carried out by [Hao et al, 2009].

#### *3.4.4 Analyzing Users Social Network*

Social networks are very helpful for determining the trustworthiness of outsiders and hence recent spam filtering approaches have started to exploit social network interactions to distinguish between spam and ham. For their social network based classification scheme, [Boykin and Roychowdhury, 2005] analyzed the e-mail header fields to construct a social network graph of the user, and then classified e-mail messages based on 'clustering coefficient' of the graph subcomponent. The clustering coefficient is *very low* for spammers, while it is *high* for a network of friends. The algorithm was found to be immune to false positives and could correctly classify 50% of all e-mails correctly. [Chirita et al, 2009] and [Golbeck and Hendler, 2004] further developed the idea of creating a social network graph for inferring reputation ratings of individuals or e-mail addresses.

Table 2, summarizes and categorizes popular machine learning attempts by authors according to perspective (Algorithm, Architecture, Methods, and Trends). Articles classified under '*Algorithm*' reflect research that essentially focused on classification algorithms and their implementations and evaluations. Articles classified under '*Architecture*' concentrated on work mainly involved with the development of spam filtering infrastructures. Articles classified under '*Methods*' refers to study of the existing filtering methods while '*Trends*' speaks of discourses concentrating on emerging methods and the adaptation of spam filtering methods over time. Limitations listed in the last column, corresponding to each article are as acknowledged by the authors themselves.

## 4 Methods for Mitigating E-mail Spam

Although there are '*social*' methods like legal measures and personal measures (*e.g.* never respond to spam, never forward chain-letters) to fight spam, they have had a narrow effect on spam so far is seen by the number of spam messages received daily by users. Technical measures seem to be the most effective in countering spam. Prior to machine learning techniques, many different technical measures were employed for spam filtering, like - rule-based spam filtering, white lists, black lists, challenge-response (C/R) systems, spam filtering, honey pots, OCR filters, and many others, each with its own merits and drawbacks. Black-lists, white-lists, challenge-response (C/R) systems, etc. are origin-based techniques used by reputation-based filters. We discuss briefly some of these popular approaches:

### 4.1 Heuristic Filters

Initial spam filters followed the '*knowledge engineering*' approach and were based on coded rules or heuristics Sanz [2008]. A content-based heuristic filter analyzes the contents of a message M and classifies it to spam or ham based on the occurrence of 'spammy' words like '*viagra*' or '*lottery*' in it. They were designed based on the knowledge of regularities or patterns observed in messages Guzella and Caminhas [2009]. Cohen's Cohen [1996] was one of the earliest attempts to use learning machines that classify e-mail. Based on the RIPPER rule-learning algorithm he employed a new method for learning from corpus sets of "*keyword-spotting rules*" to classify personal e-mails into pre-defined categories. He showed that the RIPPER algorithm can achieve a comparable performance to a traditional information retrieval (IR) method based on TF-IDF weighting.

The drawback of heuristic filters is that maintaining an effective set of rules is a time consuming affair, moreover the rules have to be constantly updated to keep up with the newest trends in spam. Spammers began employing content "obfuscation" (or obscuring), by disguising certain terms that are very common in Spam messages (e.g., by writing "v!@gra" instead of "*viagra*", or "F*r*e*e" instead of "*Free*") on an attempt to prevent the correct identification of these terms by Spam filters. Moreover writing regular expression-based rules are hard and error prone. In spite of these limitations, Symantec Brightmail Sanz [2008], a rule-based filter solution was a success from 2004 till the end of the last decade. It could even track down IP addresses that sent mostly junk mail and performed competitively to *SpamBayes* - a popular Nave Bayes-based anti-spam solution.

**Table 2** A summary popular machine learning attempts by authors according to perspective (Algorithm, Architecture, Methods, and Trends), with their strengths and limitations.

| Ref. | Perspective | Strengths and Limitations |
|---|---|---|
| Tretyakov [2004] | Naive Bayes, k-NN, ANN, SVM | Techniques benefits beginners. |
| | Algorithms, Methods | *Does not* deal with feature selection. |
| [Androutsopoulos et al, 2006] | Naive Bayes, LogitBoost, SVM | Resulted in - *LingSpam* and *PU1.* |
| | Algorithm, Methods, Trends | *Ignored* headers, HTML, attachments. |
| [Carpinter and Hunt, 2006] | Bayesian filtering | Broad review of implementations. |
| | Methods, Architecture | *Focuses primarily* on automated, filters. |
| [Blanzieri and Bryl, 2008] | SVM, TF-IDF, Boosting | Explains feature extraction methods. |
| | Algorithms, Methods, Trends | *Does not cover* neighboring topics. |
| [Cormack, 2008] | SVM, Perceptron, Winnow, OSBF | Testing achieves FPR = 0.2 %. |
| | Algorithms, Methods, Trends | User feedback *difficult to simulate.* |
| [Guzella and Caminhas, 2009] | Regression, Ensembles | Focuses on textual and image analysis. |
| | Algorithms, Methods | Focuses *only* on application specific aspects. |
| [Almeida and Yamakami, 2010] | SVM, Naive Bayes | Proposed Matthews correlation coefficient (MCC). |
| | Algorithms, Methods | Need to compare with other algorithms & corpuses. |
| [Almeida and Yamakami, 2012] | MDL principle, SVM | Uses six, well known, large public databases. |
| | Algorithms, Methods | Bogofilter, SpamAssassin filters *not considered.* |
| [Caruana and Li, 2012] | Signature, k-NN, ANN, SVM | Focuses on distributed computing paradigms. |
| | Methods, Architecture | *Avoids* implementation and interoperability issues. |
| [Wang et al, 2013] | Statistical analysis, n-grams | Investigated *topic drift.* |
| | Trends | Limited datasets. |

## 4.2 Blacklisting

A blacklist of E-mail addresses or IP addresses of the server from which spam is found to originate is created and maintained either at the user or server level. If a user receives an e-mail from any of these addresses, the message is automatically blocked at the SMTP connection phase. This method requires only a simple lookup in the blacklist every time; hence the computational cost is low. Black-lists include *Real-time Blackhole ListS* (RBL) and *Domain Name System Black-lists.* Common black-list databases include proxies or open relays, networks or individual addresses guilty of sending spam. Google blacklists and SpamHaus [7] are examples of blacklists.

Blacklist techniques though effective, suffer from many drawbacks. A legitimate address may be blacklisted by the filter erroneously or arbitrarily. Innocent users can get victimized and entire domains (e.g. Hotmail) can get blocked when e-mail IDs or IP addresses are used by spammers without the owners consent. As spam-mers resort to use of new E-mail IDs or IP addresses to bypass blacklists, frequent updates are required to keep the blacklists up-to-date. Lately, use of botnets by spammers creates an extremely huge number of IP addresses to be blacklisted. While the time and effort for updating can be overwhelming, any lag in its timeliness leads to vulnerabilities.

[7] http://www.spamhaus.org/

### 4.3 Whitelisting

Whitelisting is the reverse of blacklisting. An e-mail whitelist is a list of pre-approved or trusted contacts, domains, or IP addresses that are able to communicate to a mail user. All e-mails from fresh e-mail addresses are blocked by this method. This restrictive method may introduce an extremely high false positive rate instead of reducing it. Such a method may be good for instant messaging environments but is not a good choice as it prohibits establishing new contacts through e-mail. Moreover if spammers somehow got their hands on the whitelist, it would be easy to evade the filter using spoofed addresses, or using well-known whitelisted mailing lists. This method requires a lot of maintenance but provides moderate filtering rate. It can be employed together with other anti-spam techniques [Michelakis et al, 2004].

### 4.4 Greylisting

When an SMTP client connects requesting for a session for the first time, the recipient server may check if the IP address of the sender or its e-mail address is blocked or pre-approved. It may happen they are neither in the blacklist nor in the whitelist. In that case the message is rejected temporarily and the recipient MTA responds with an SMTP temporary error message. The recipient MTA then records the identity of recent attempts and its databases are updated with the new clients information; as required by SMTP RFC [P. Resnick, 2001], the client retries at a later point of time. The next attempt may be accepted for legitimate senders. This method assumes that spammers do not waste time in queuing or retrying their messages and those who do so will probably end up being blacklisted in public blacklists (DNSBLS) during the two attempts. While this technique seems very effective, evading it can also be very simple. The spammers can use zombies to do the work of retrying for the spammer.

### 4.5 Challenge Response (CR) systems

While white-lists place the burden of determining the authenticity of contacts on the receiver, Challenge-Response (CR) systems transfer the burden of authentication back to the sender. After sending an e-mail to the receiver, the sender receives a challenge from the receiving Mail Transfer Agent (MTA). The challenge may range from a simple question to a CAPTCHA (”*Completely Automated Public Turing test to tell Computers and Humans Apart*”). The sender is obliged to reply correctly in his response; else his message will be deleted or put into spam folder. While this method is effective in catching spam from automated systems or botnets, it introduces an undesirable delay in the delivery process. CR systems are controversial solutions and are often criticized due to this inconvenience caused by the overhead in communication. Besides legitimate e-mails from automated mailing lists may also be blocked since these will fail the challenge. In addition, CR systems are also believed to be the cause behind the backscatter e-mail phenomenon. [Isacenkova and Balzarotti, 2011] developed a real world deployment of a CR based anti-spam system and evaluated its effectiveness and impact on end-users.

### 4.6 Collaborative Spam filtering

Spammers typically send spam to a vast number of recipients. It is likely that the same spam has been received by somebody else. Collaborative spam filtering is a distributed approach to filtering spam where a whole community works together with a shared knowledge about spam [Méndez et al, 2006], [Sophos, 2013], and [Garriss et al, 2006]. The collaborative approach does not consider the content of e-mails; rather it requires the accumulation of any identifying information concerning spam messages, like - the subject, sender, the result of computing a mathematical function over the email body, etc. Spam messages have digital footprints which are shared with the community by early receivers. The community users then use these spam fingerprints for identifying spam e-mails. Vipuls Razor [8], Pyzor [9], DCC (Distributed Checksum Clearinghouse) [10] are examples of collaborative spam filters on the web. Though it is certain that collaborative techniques show great promise, however such schemes suffer from scalability issues and some underlying implicit assumptions.

[8] http://razor.sourceforge.net/
[9] https://github.com/SpamExperts/pyzor
[10] http://www.rhyolite.com/dcc/

### 4.7 Honey pots

A honeypot is a decoy server or system set up solely to collect spam or gather information about intruders Andreolini et al [2005]. It is also used to identify e-mail address harvesters with the help of specially generated e-mail addresses and to detect e-mail relays. It is a finger-print based technique for content based spam filtering. Honeypots do help security professionals and researchers learn the techniques used by attackers to compromise computer systems. Bringer et al [2012] present a proper survey on the evolution in honeypots as well as advances and the current trends in honeypot research to cope with recently emerging security threats.

### 4.8 Signature Schemes

Most current antivirus products work on the basis of signatures. The hashes of previously identified spam messages are kept in a database at the Mail Transfer Agent (MTA) level. All incoming e-mails are checked against these hashes to distinguish between spam and legitimate e-mails. Because signatures match exact patterns, this scheme can detect known spams with a very high level of confidence. However, a strong shortcoming is that unknown or newly generated spam will be able to get past this filter without being detected. Signature databases need to be updated hourly, daily or weekly. The database can swell as thousands of spams are generated every day. Spammers can introduce a random string into spam mails to generate different hashes.

This review article examined a number of major earlier surveys on spam filtering over the period (2004-2015). Perusing the different spam techniques and the methods used by researchers to combat spam, taxonomy of spam filtering techniques is presented above (Table 3).

## 5 Machine Learning Approach to E-mail Spam filtering: The Algorithms

Spam filtering is a binary classification task, in which *legitimate* (good or *ham*) e-mails are treated as negative (-) instances, and *spam* as positive (+) instances [Song et al, 2009]. Machine Learning is a subfield of computer science that explores the design and development of computer systems that automatically improve their performance in a task based on experience. Automatic e-mail classification uses statistical approaches or machine learning techniques and aims at building a model or a classifier specifically for the task of filtering spam from a users mail stream. Some of the most popular Machine Learning techniques to counter spam filtering are Naive Bayes [M. Sahami, S. Dumais, D. Heckerman et al, 1998], [Androutsopoulos et al, 2000a], Support Vector Machines [] Woitaszek and Shaaban [2003], [Amayri and Bouguila, 2010]], Decision Trees [Yeh et al, 2005], [Toolan and Carthy, 2010], Neural Networks [Wu, 2009], [Soranamageswari and Meena, 2010], etc. The building of the model or classifier requires a set of pre-classified documents (*training set* or an initial corpus). The process of building the model is called *training*.

Machine learning algorithms have achieved more success among all previous techniques (discussed in Section 4) employed in the task of spam filtering [Fdez-Riverola et al, 2007a], [Almeida et al, 2010]. In fact, the success stories of Gmail [Taylor et al, 2007], [The, 2010], can be ascribed to their timely transition and successful use of Machine Learning for filtering not just incoming spam but other abuses like Denial-of-Service (DoS), virus delivery, and other imaginative attacks [Taylor et al, 2007]. Today the most successful spam filters are based upon the statistical foundations of Machine Learning. In part it is because it is easier to train and build a classifier on e-mails that individual mail users receive, than to build and tune a set of filtering rules. Machine Learning based spam filters also retrain themselves while put in use and minimizes manual effort while delivering superior filtering accuracy. In this section we explore the underlying theory and aim to present a clear picture of popular Machine Learning algorithms employed in spam filtering for the benefit of readers unfamiliar with them. Table 3 provides a taxonomy of e-mail spam filtering techniques.

### 5.1 Naive Bayes (NB)

*Naive Bayes* classifiers are a technique that has remained popular over the years and are arguably the most well-known statistical spam classifier. It is called '*naive*' because it ignores possible dependencies or correlations among inputs and reduces a multivariate problem to a group of *uni-variate* problems [M. Sahami, S. Dumais, D. Heckerman et al, 1998]. It employs a probabilistic approach to inference. It does not need any complicated iterative parameter estimation schemes, as in *Discriminant analysis*. It is easy to construct, easy to interpret, surprisingly effective and can be readily applied to huge data sets [Wu et al, 2007], making it extremely popular among users. Bayesian methods typically require prior knowledge of many probabilities *e.g.*, according to Grahams [Graham, 2002a] corpus, the word '*sex*' indicates a 97% probability that the containing e-mail is spam. Similarly, words like '*viagra*' and '*refinance*' will have high spam probability values, while names of friends and siblings will have low spam probability values. These *apriori* probabilities are combined with the observed data set - which is a sizeable collection of e-mails that has already been categorized as '*spam*' and '*ham*', to determine the final probability that an e-mail message is either spam or legitimate. Even with the flawed assumption of presumed decorrelation, Bayesian classifiers work extremely well and are surprisingly effective [M. Sahami, S. Dumais, D. Heckerman et al, 1998], [Pantel and Lin, 1998], [Graham, 2003] and [Yerazunis, 2004].

**Table 3** A taxonomy of e-mail spam filtering techniques.

| Reputation-based | Content-based (Textual) | Content-based (Multimedia) |
|---|---|---|
| `Reputation based`<br>└ `Origin based`<br>  └ `Blacklists`<br>  └ `Whitelists`<br>  └ `Origin Diversity Analysis`<br>└ `Social Networks`<br>  └ `Implicit`<br>  └ `Explicit`<br>└ `Traffic analysis`<br>  └ `Mail Volume`<br>  └ `SMTP Flow`<br>└ `Protocol based`<br>  └ `C-R Systems`<br>  └ `Greylisting` | `Textual content`<br>└ `Heuristics`<br>  └ `Rule based`<br>└ `Fingerprint based`<br>  └ `Honeypots`<br>  └ `Digest based`<br>  └ `Signature/Checksum schemes`<br>└ `Machine Learning`<br>  └ `Naive Bayes`<br>  └ `Support Vector Machines`<br>  └ `Decision Trees`<br>  └ `Clustering`<br>  └ `Ensembles` | `Multimedia content`<br>└ `OCR techniques`<br>  └ `Keyword detection`<br>  └ `Text Categorization`<br>  └ `High Level Analysis`<br>└ `Low-level Features`<br>  └ `Image Classification`<br>  └ `Near Duplicate Detection` |

Nave Bayes method has become extremely popular due to the high levels of accuracy that it can potentially provide and it often serves as a baseline classifier for comparison with other filtering approaches. Bayesian filters are the most employed filters for classifying spam nowadays Guzella and Caminhas [2009], Metsis et al [2006] and can operate either on the network mail server level or on client e-mail programs.

One limitation of standard Bayesian filters is that it ignores the correlation among inputs or events; i.e. such filters do not consider that the words '*special*' and '*offers*' are more likely to appear together in spam e-mail than in legitimate e-mail Carpinter and Hunt [2006]. But text analysis confirms that words have a very significant correlation and are not chosen randomly. In spite of this over simplistic assumption, Bayesian classifiers have been found to work remarkably well Androutsopoulos et al [2006] and Almeida et al [2010]. However to address this limitation, Yerazunis [2003] and Siefkes et al [2004] introduced sparse binary polynomial hashing (SBPH) and orthogonal sparse bigrams (OSB). SBPH is a generalization of Bayesian filtering that can match mutating phrases as well as individual words or tokens, and uses the Bayesian Chain Rule (BCR) to combine the individual feature conditional probabilities into an overall probability. SBPH had a more expressive feature space and delivered ¿99.9% accuracy on real-time e-mail without white-lists or blacklists from as little as 500K of pre-categorized text. However, SBPH was computationally expensive; OSB retains the expressivity of SBPH but avoids most of the cost. A filter based on OSB, along with the non-probabilistic Winnow algorithm as a replacement for the Bayesian Chain Rule outperformed SBPH by 0.04% error rate; however, OSB used just 6, 00,000 features, while SBPH used 1,600,000 features to reach best results. Yerazunis [2004] argued that most Bayesian filters seem to reach a plateau of accuracy at 99.9 percent so enhancements were necessary. They set up a SBPH/BCR classifier and compared three different training methods: TEFT  Train Every Thing, TOE  Train Only Errors, TUNE  Train Until No Errors, and found TOE training to be acceptable in performance and accuracy. Different extensions to Bayesian filtering as Token Grab Bag, Token Sequence Sensitive, Sparse Binary Polynomial Hashing with Bayesian Chain Rule (SBPH/BCR), Peaking Sparse Binary Polynomial Hashing, Markovian matching, were also tested. Markovian matching produced the best performance of all the filters.

According to Ludlow [2002], the vast majority of the tens of millions of spam e-mails might be the handiwork of only 150 spammers around the world; Again, authors have '*textual fingerprints*', at least for texts produced by writers who are not consciously changing their style of writing across texts, as argued by Baayen et al [2002]. Therefore authorship identification techniques can be used to identify the '*textual fingerprints*' of this small group and eliminate a significant proportion of spam. Brien and Vogel [2003] were the first to apply authorship identification techniques as '*Chi by degrees of freedom*' method to the area of e-mail spam filtering. The authors examined the Nave Bayesian method in relation to this authorship identification technique. They found that the Bayesian method was very effective when characters were used as tokens, rather than when words were used as tokens. The '*Chi by degrees of freedom*' method when used with characters as tokens had an error rate lesser than the Bayes method. They concluded that tokens chosen affected the precision and recall parameters. Taking a leaf out of text classification, Song et al [2009] proposed a correlation-based document term weighting method to address the problem of low-FPR classification in the context of Nave Bayes.

[Chih-Chin Lai and Tsai, 2004] conducted systematized experiments on e-mail categorization involving Naive Bayes (NB), Term Frequency - Inverse Document Frequency (TF-IDF), k-Nearest Neighbor (k-NN), and Support Vector Machines (SVMs). NB, TF-IDF and SVM achieved satisfactory results while k-NN had the worst performance out of all. It was seen that *stemming* did not affect performance, however employing *stopping* procedure yielded better performance. They concluded that combining the different techniques seemed a very promising prospect. [Lai, 2007] has made a similar comparative study on three commonly used algorithms in Machine Learning NB, k-NN and SVMs. From experimental results, NB and SVM were found to perform better than k-NN. [Youn and Mcleod, 2006] and [Yu and Xu, 2008] noted similar experiments with four machine learning algorithm each. [Seewald, 2007] investigated the simple Naive Bayes learner represented by *SpamBayes*, and two variants of Naive Bayes learning, *SA-Train and CRM-114*. SA-Train incorporated background knowledge made up of rules while CRM-114 considered multi-word phrases and their probability estimates. It was seen that all three systems performed equally well and the addition of background knowledge to SA-Train and the extended description language in the case of CRM-114 considering multi-word phrases failed to improve Bayesian learning significantly. SpamBayes offered the most stable performance and deteriorated least over time. [Almeida et al, 2010] reported that probabilistic approaches like Bayesian classification suffer from the '*curse of dimensionality*'. They verified how dimensionality reduction influences the accuracy of Nave Bayesian spam filters.

### 5.2 Support Vector Machines (SVMs)

Support vector machines (SVMs) are ranked as one of the best '*off-the-shelf*' supervised learning algorithm. SVMs have become one of the most sought-after classifiers in the Machine Learning community because they provide superior generalization performance, require less examples for training, and can tackle high-dimensional data with the help of kernels [Rios and Zha, 2004] [Wu et al, 2007]. Support vector machines (SVMs) result by mapping the feature vectors (training data) into a linear or non-linear feature space through a kernel function. The feature space generates an optimal separating hyper-plane (OSH) which splits the positive samples and the negative samples with maximum margin. The hyper-plane is then employed as a non-linear decision boundary for use in real-world data.

[Drucker et al, 1999] used SVMs for content-based classification and equated their performance with other classifiers - Ripper, Rocchio, and boosting of C4.5 decision trees. It was found that boosting trees and SVMs attained good performance with regard to speed and accuracy during testing. SVMs with binary features produced best results, required lesser training, and their performance did not degrade when too many features were used. Woitaszek and Shaaban [Woitaszek and Shaaban, 2003] utilized an SVM-based filter for Microsoft Outlook to identify commercial e-mail. Classification models for spam and ham messages were built by the SVM using personal and impersonal dictionaries. Both yielded identical results attaining a best accuracy of 96.69%. [Rios and Zha, 2004] experimented with SVMs and Random Forests (RFs) and compared them against Naive Bayes models. They concluded that SVM and RF classifiers were equivalent, and that the RF classifier had greater robustness at low false positive (FP) rates; they both outperformed Naive Bayes models at low FP rates. [Tseng and Chen, 2009] proposed a complete spam detection system MailNET, which is an incremental SVM model on dynamic e-mail social networks. Although SVMs provide high accuracy for spam filtering, they have been generally associated with high computational cost and some expensive false positive errors, hence, few solutions were offered, *e.g.* Online SVMs [Sculley and Wachman, 2007], Ensemble of SVMs [Blanco et al, 2007], etc. A detailed study of various distance-based kernels and spam filtering behaviors employing SVM is found in [Amayri and Bouguila, 2010].

### 5.3 Clustering Techniques

Clustering is the task of grouping a set of patterns into similar groups. Clustering techniques have been widely studied and used in a variety of application domains. Spam filtering datasets often have true labels available and clustering algorithms, being unsupervised learning tools are not always closely related with true labelings. However given suitable representations, most clustering algorithms can partion e-mail spam datasets into ham and spam clusters. This was demonstrated by [Whissell and Clarke, 2011] in a novel investigation of e-mail spam clustering. The results were surprisingly significant as their clustering based approach bettered those of previously published state-of-the-art semi-supervised approaches, hence proving that clustering can be a powerful tool for e-mail spam filtering.

Prior to this, Sasaki and Shinnou [Sasaki and Shinnou, 2005] had proposed spam detection technique making use of the text clustering through a vector space model. [Basavaraju and Prabhakar, 2010] presented an effective clustering algorithm integrating K-means and BIRCH algorithm features. K-means algorithm worked well for small scale data sets. BIRCH with K-Nearest Neighbour Classifier (K-NNC) was found to be the ideal combination as it performed better with large data sets. [Debarr and Wechsler, 2009] relied on using term frequency and inverse document frequency representation for e-mails and employed the Partitioning Around Medoids (PAM) clustering algorithm to cluster a uniform sample of 25% of messages in the training pool. Clustering combined with Random Forests for classification and active learning for refinement produced the best Area Under Curve (AUC) of 95.2%. These works conclude that employing the ham/spam clusters is a effective method for spam detection and because a ham/spam split is a natural clustering for an e-mail spam dataset, clustering techniques should be investigated further as a tool for more robust content based spam filters.

### 5.4 Ensemble Classifiers

Ensemble learning is a novel technique where a set of individual classifiers are trained and brought together to enhance the classification accuracy of the overall system on the same problem (spam detection). An ensemble of classifiers is very effective for classification tasks and offers good generalization. Spam filters have to deal with a diversity of spams, so it needs to continually evolve in order to detect new types of spam (future spam), and at the same time not allow 'classical' spam to evade the filter. Therefore, [Guerra et al, 2010] had suggested that combining old and new filters (*e.g.* using ensemble classifiers) may be an interesting strategy to deal with the diversity of spams. The most popular ensemble classifiers are *bagging* and *boosting*.

Bagging (or *bootstrap aggregating*) is an ensemble meta-learning algorithm that is usually applied to decision tree methods, *e.g.* Random Forest algorithm is an ensemble technique for decision trees that is known to achieve very high classification accuracy. [Biggio et al, 2011] employed bagging ensembles to exploit against poisoning attacks on spam filters. Random forests have also been used in the spam detection model described in [Debarr and Wechsler, 2009] and [Lee et al, 2010b].

Boosting [Biggio et al, 2011] involves algorithms that build a single strong learner from a set of weak learners. AdaBoost is the most common implementation of Boosting. Boosting for filtering of spam messages was first reported by [Carreras and Marquez, 2001]. [Androutsopoulos et al, 2006] compared four most promising learning algorithms from earlier work - LogitBoost, Nave Bayes, Flexible Bayes and linear SVM. The authors studied the role of attributes characterizing *n-grams* frequencies and explored the effect of attribute size and training set in a cost-sensitive framework context. Using evaluation measures as in [Androutsopoulos et al, 2000b], and the PU1 corpus in experiments, [Carreras and Marquez, 2001] proved the definite effect of boosting in decision-tree filters. Methods based on boosting outperformed Naive Bayes and Decision Trees algorithms when tested on the PU1 corpus. [Sakkis et al, 2001] experimented with combining a memory-based classifier with a Naive Bayes filter with another memory-based classifier as president in a stacking framework. They achieved impressive *precision* and *recall* and concluded that stacking consistently raises the performance of the overall filter. He and Thiesson [He and Thiesson, 2007] proposed a new asymmetric boosting method - Boosting with Different Costs and applied it to spam filtering. [Neumayer, 2006], [Shi et al, 2012], and [Blanco et al, 2007] also discuss the application of an ensemble learning to spam filtering.

## 6 Evaluation Measures and Benchmarks

Ideally spam filters should be evaluated on large, publicly available spam and ham databases. Sometimes *Accuracy* (Acc), the ratio of messages correctly classifies is used as an integrated measure for performance. If $N_L$ and and $N_S$ signify the number of legitimate messages and spam messages to be classified, then we define *Accuracy* (Acc) and *Error* (Err) of the spam filter as -

$$Acc = \frac{|L \to L| + |S \to S|}{N_L + N_S} \text{ and } Err = 1 \text{ - } Acc = \frac{|L \to S| + |S \to L|}{N_L + N_S}$$

Accuracy and Error consider both False Positive $|L \to S|$ and False Negative $|S \to L|$ events to carry equal cost. However, spam filtering involves asymmetric error costs. Failing to identify a ham, *i.e.* misclassifying a ham as spam (a *False Positive event*) is generally a costlier mistake than missing a spam (a *False Negative event*). For *e.g.* A business letter from the boss or a personal message from a spouse quarantined (and delayed) or deleted can lead to serious consequences, while seeing a spam in our inbox may cause just a slight irritation. True Positive event $|L \rightarrow L|$ is when a ham e-mail is correctly classified as ham. True Negative event $|S \rightarrow S|$ is when a spam e-mail is correctly classified as spam. With this in mind, the *False Positive Rate* (FPR) - the proportion of legitimate e-mails identified as spam is represented as -

$$FPR = \frac{\#ofFalsePositives}{\#ofFalsePositives+\#ofTrueNegatives}$$

Again, failing to identify spam *e.g.* e-mails containing viruses, worms, or phishing baits as payload can incur significant risks to the user. *False Negative Rate* (FNR) *i.e.* the proportion of spam messages that were classified as legitimate, is another suitable measure.

$$FNR = \frac{\#ofFalseNegatives}{\#ofTruePositives+\#ofFalseNegatives}$$

Superior spam classifiers have lower FPR and FNR. The two-dimensional quantity (FNR, FPR) denotes the effectiveness of hard classifiers while the effectiveness of soft classifiers may be denoted by a set of such pairs defining a curve - an ROC (*Receiver Operating Characteristics*) curve. ROC analysis are an excellent performance metric in spam filtering. A spam filter whose ROC curve strictly lies above that of another is the better filter in all deployment scenarios. [Cormack, 2008].

Two measures borrowed from Information Retrieval '*Recall*' and '*Precision*' are often used for capturing the effectiveness and quality of spam filters respectively [Androutsopoulos et al, 2000a]. If $|S \rightarrow L|$ signifies the number of spam messages classified as legitimate, and $|S \rightarrow S|$ signifies the number of legtimate messages classified as spam respectively, and likewise for $|L \rightarrow L|$ and $|L \rightarrow S|$ then *Spam Recall* ($R_s$) and *Spam Precision* ($P_s$) are defined by the equations:

$$R_s = \frac{|S\rightarrow S|}{|S\rightarrow S|+|S\rightarrow L|} \text{ and } P_s = \frac{|S\rightarrow S|}{|S\rightarrow S|+|L\rightarrow S|}$$

Recall ($R_s$) is a measure of the number of spam messages successfully blocked by the filter (i.e. its *effectiveness*), while Precision ($P_s$) measures the number of the messages classified as spam by the filter that were indeed spam (i.e. its *quality* or *safety*) [Androutsopoulos et al, 2006] [Sakkis et al, 2001]. Comparing spam filters based on ($R_s$) and ($P_s$) is tricky despite with each configuration giving ($R_s$) and ($P_s$) values.

False positives are considerably more expensive ($\lambda$ times) when compared with false negatives [Androutsopoulos et al, 2000a] [Androutsopoulos et al, 2006]. Here, $\lambda$ is a parameter that specifies how 'dangerous' or 'costly' it is to misclassify legitimate e-mail as spam and reflects the extra effort it requires from the user to recover from failures of the filter. For many users false positives are unacceptable. [Androutsopoulos et al, 2006] suggested this cost sensitivity be taken into account by treating each legitimate message to be equal to $\lambda$ messages. Cost-sensitive measures *Weighted Accuracy* (WAcc), *Weighted Error Rate* (WErr) and *Total Cost Ratio* (TCR) [Clark, 2008] are used as shown in the formula.

$$WAcc = \frac{\lambda|L\rightarrow L|+|S\rightarrow S|}{N_L+N_S} \text{ and } WErr = 1 - WAcc = \frac{\lambda|L\rightarrow S|+|S\rightarrow L|}{N_L+N_S}$$

$$TCR = \frac{N_S}{\lambda|L\rightarrow S|+|S\rightarrow L|}$$

The Total Cost Ratio is used to compare the effectiveness of a filter for a given $\lambda$ when compared with a baseline setting [Guzella and Caminhas, 2009]. It is an evidence of the improvement brought about by the filter. This cost-sensitive evaluation uses the $\lambda$ parameter to adjust the weight of a false positive. There are three values for $\lambda$ used commonly in spam literature, $\lambda = 1$, 9, 999 [Androutsopoulos et al, 2000b], [Androutsopoulos et al, 2000a], [Androutsopoulos et al, 2006], [Sakkis et al, 2001] and [Clark, 2008]. These values represent the situations when a false positive equals a false negative, or a false positive is 9 times a costlier mistake than a false negative, or 999 times costlier. Greater TCR values indicate superior performance. *F-measure* or *F-score* is another combining measure that combines both *Precision* ($P_s$) and *Recall* ($R_s$) metrics in one equation. It can be interpreted as the weighted harmonic mean of both.

$$F-measure = \frac{2.Precision.Recall}{Precision+Recall}$$

### 6.1 Publicly Available Benchmark Datasets

Most of the datasets publicly available are static datasets with very few concept drift datasets. Many authors construct their own image spam or phishing corpus. Table 4 below lists public corpora with associated information used in spam filtering experiments.

**Table 4** Publicly available benchmark datasets on E-mail Spam.

| Corpus Name | Number of Messages/Images | | Spam Rate | Year of Creation | Reference/Used |
|---|---|---|---|---|---|
| | Spam | Ham | | | |
| SpamAssassin | 1897 | 4150 | 31% | 2002 | [Méndez et al, 2006] |
| Enron-Spam | 13,496 | 16,545 | - | 2006 | [Koprinska et al, 2007] |
| LingSpam | 481 | 2412 | 17% | 2000 | [Sakkis et al, 2001] |
| PU1 | 481 | 618 | 44% | 2000 | [Attar et al, 2011] |
| PU2 | 142 | 579 | 20% | 2003 | [Zhang et al, 2004] |
| PU3 | 1826 | 2313 | 44% | 2003 | [Zhang et al, 2004] |
| PUA | 571 | 571 | 50% | 2003 | [Zhang et al, 2004] |
| Gen Spam | 41,404 | | 78% | 2005 | [Cormack and Lynam, 2007] |
| Spambase | 1813 | 2788 | 39% | 1999 | [Sakkis et al, 2001] |
| ZH1 | 1205 | 428 | 74% | 2004 | [Zhang et al, 2004] |
| TREC 2005 | 52,790 | 39,399 | - | 2005 | [Androutsopoulos et al, 2000a] |
| TREC 2006 | 24,912 | 12,910 | - | 2006 | [Androutsopoulos et al, 2000c] |
| TREC 2007 | 50,199 | 25,220 | - | 2007 | [Debarr and Wechsler, 2009] |
| Spam Archive | >2,20,000 | | 100% | 1998 | [Almeida and Yamakami, 2012] |
| Biggio | 8549 | 0 | - | 2005 | [Biggio et al, 2006] |
| Princeton Spam Image Benchmark | 1071 | 0 | - | - | [Biggio et al, 2006] |
| SpamArchive | >2,20,000 | | 100% | 1998 | [Almeida and Yamakami, 2012] |
| Dredze Image Spam Dataset | 3927 | 2006 | - | 2007 | [Almeida and Yamakami, 2012] |
| Phishing Corpus | 415 | 0 | - | 2005 | [Abu-nimeh et al, 2007] |

## 7 Future Challenges and Conclusion

Spam filtering is an 'arms race' marked by an increase in the sophistication in spam construction techniques as well as spam filtering techniques [Goodman et al, 2007]. Characterization and measurement studies have been developed in content-based spam filtering [Pu and Webb, 2006], [Blanzieri and Bryl, 2008]. The evolution of the infrastructure used by spammers to disseminate spams over the network is seen in their migration from simple obfuscation techniques, to image spam and to compromised machines. The dynamic nature of spam and the reactivity of spammers make e-mail spam filtering an active research area. E-mail spam filtering will remain a persistent problem and some of the most interesting challenges in the future of e-mail spam filtering could be -

### 7.1 Handling Concept Drift

In the actual world, concepts change over time in unanticipated ways and are therfore hard to predict. Changes in the statistical properties of context can lead to a change in the target variable or concept. Concept drift is distinguished in literature as 'sudden' and 'gradual' [Tsymbal et al, 2008]. For *e.g.*, a student graduating from college might all of a sudden develop financial concerns, whereas, in a biomedical context, pathogen sensitivity may gradually evolve with the passage of time as bacterial pathogens develop immunity to antibiotics that used to be effective earlier. Hidden changes in context affects not just the target concept but also causes an alteration in the underlying data distributions [Delany et al, 2005], making the learning task increasingly complicated and requiring special approaches. Models built on old data become less accurate or inconsistent making the rebuilding of the model imperative (called *virtual concept drift*). Spam filtering is a dynamic problem that involves concept drift. While the understanding of an unwanted message may remain the same, the statistical properties of the spam e-mail changes over time since it is driven by spammers involved in a never-ending arms race with spam filters. Another reason for concept drift could be the different products or scams driven by spam that tend to become popular. The dynamic nature of spam is one of its most testing aspects. An effective spam filter must be able to track target concept drift and swiftly adapt to it. Research on concept drift confirms lazy learning techniques to be the most effective models against concept drift [Tsymbal, 2004], [Tsymbal et al, 2008]. Most of the earlier evaluations did not try to deal with concept drift, or with real-world datasets that have some concept drift. Few authors tried to address concept drift in spam filtering using Case-Base Reasoning [Delany et al, 2005], Instance-Based Reasoning [Fdez-Riverola et al, 2007b], Ensemble Learning [Tsymbal et al, 2008], Language Model technique [Hayat et al, 2010]. A particular challenge in handling concept drift is in distinguishing between true concept drift and noise. Research in concept drift is a very active area in spam filtering.

### 7.2 Eliminating False Positives

Spam filtering is often viewed as a straight text categorization problem. But e-mail is not just text, it also has structure, hence in reality it turns out to be a more complicated problem than straightforward classification. One complication arises from the cost-sensitivity associated with the spam filtering problem. The cost of inadvertently restricting a ham message is more than that of a spam message evading the filter (see section 6). Such mislabeling of e-mail is completely unacceptable to users as it can lead to loss of important information or even more serious consequences. Moreover, in this case the user has to review the messages sorted out to the spam folder and it somehow defeats the whole purpose of spam filtering [Tretyakov, 2004]. Content-based spam filtering systems, though widely adopted as a successful spam defense strategy, has unfortunately substituted the spam issue with a false positive one. Such systems achieve a high accuracy but there exists some false positive tradeoff. False positives are *more severe and expensive* than spam. Although significant attempts *e.g.* Reliable e-mail [Garriss et al, 2006] have been made, nevertheless, to make e-mail reliable, spam filters must reduce the incidences of false positives. Reduction of false positives is another domain in e-mail spam analysis where much work needs to be been done on leveraging existing algorithms.

### 7.3 Emerging Spam Threats

One of the biggest spam problems today even as spam e-mail volumes associated with botnets are receding is the *snowshoe spam*. Showshoe spamming is a technique that uses multiple IP addresses, websites and sub-networks to send spam, so as to avoid detection by spam filters. The term '*snowshoe*' spam describes how some spammers distribute their load across a larger surface to keep from sinking, just as snowshoe wearers do [McAfee, 2012] [Sophos, 2013]. Social networks have also become a hunting ground for spammers. With many users migrating to social networks as a means of communication, spammers are diversifying in order to stay in business. The personal information revealed in social networks is gleaned by spammers to target unsuspecting victims with tailored e-mails.

### 7.4 Prioritising E-mails

E-mail prioritization is an urgent research area with not much research done. In addition to basic communication, our e-mails are 'overloaded' in the sense of being used for a wide variety of other tasks - communication, advertisements, reminders, contact management, task management, and cloud storage. There is a serious need to address the information overload issue by developing systems that can learn personal priorities from data and identify important e-mails for each user. Prioritizing e-mail as per its importance is another desirable characteristic in a spam filter. Prioritizing e-mail or perhaps redirecting urgent messages to handheld devices could be another way of managing e-mails [Koprinska et al, 2007]. Learning to prioritize or rank is a relatively new field in which Machine Learning algorithms are used to learn some ranking function. [Dredze et al, 2009] and [Aberdeen and Slater, 2011] are significant works on ranking algorithms for proposing useful filters that rapidly filter groups of inbox messages and search messages more easily. However importance ranking is harder than it seems as often users disagree on what is important, requiring a high degree of personalization. The result is the growth of one of the most challenging research areas in Machine Learning *i.e.* Personalized e-mail prioritization [Yang et al, 2010], which rely mostly on the analysis of social networks to model user priorities among incoming e-mail messages.

## 8 Conclusion

Future researches must address the fact that e-mail spam filtering is a co-evolutionary problem, since as the filter attempts to extend its predictive accuracy, the spammers attempt to outdo the classifiers. Hence, an effective approach should find a successful mechanism to identify the drift or evolution in spam features. Among all the traditional approaches discussed so far, the single approach that has achieved tremendous success against spam is content-based spam filtering. Fortunately, machine learning-based systems enable systems to learn and adapt to new threats, reacting to counteractive measures adopted by spammers.

No single anti-spam solution may be the right answer. A multi-faceted approach that combines legal and technical solutions and more is likely to provide a death blow to such spam. Without an effective solution spam will only continue to decrease the value of an efficient communication medium. As long as spam exists it will continue to have adverse effects on the preservation of integrity of e-mails and the user's perception on the effectiveness of spam filters. We reviewed content-based spam filtering techniques based on Machine Learning methods propounded so far, highlighting the main approaches and advancements gained by the approach. A quantitative analysis of the major reviews over the last decade was conducted. Overall the number and quality of literature demonstrates that remarkable advancements have been achieved and continue to be achieved. However some outstanding problems in e-mail spam filtering as highlighted above still remain. Till more improvements in spam filtering happen, anti-spam research will remain an active research area.

## References

(2010) An Analyst Review of Hotmail Anti-Spam Technology. A White Paper. Tech. rep., The Radicati Group Inc, URL www.radicati.com

(2014) Stanford Anti-Phishing Browser Extensions. URL https://crypto.stanford.edu/antiphishing/

Aberdeen D, Slater A (2011) The Learning Behind Gmail Priority Inbox. In: NIPS 2010 Workshop on Learning on Cores, Clusters and Clouds, pp 3–6

Abu-nimeh S, Nappa D, Wang X, Nair S (2007) A Comparison of Machine Learning Techniques for Phishing Detection. In: eCrime 07: Proceedings of the Anti-phishing Working Groups 2nd Annual eCrime Researchers Summit, New York,USA, pp 60–69

Al-jarrah O, Khater I, Al-duwairi B (2012) Identifying Potentially Useful Email Header Features for Email Spam Filtering. In: The Sixth International Conference on Digital Society, c, pp 140–145

Almeida TA, Yamakami A (2010) Content-Based Spam Filtering. In: The 2010 International Joint Conference on Neural Networks (IJCNN), Barcelona, pp 1–7

Almeida TA, Yamakami A (2012) Advances in Spam Filtering Techniques. In: Computational Intelligence for Privacy and Security, Springer Berlin Heidelberg, pp 199–214

Almeida Ta, Almeida J, Yamakami A (2010) Spam Filtering: How the Dimensionality Reduction Affects the Accuracy of Naive Bayes Classifiers. Journal of Internet Services and Applications 1(3):183–200, DOI 10.1007/s13174-010-0014-7, URL http://www.springerlink.com/index/10.1007/s13174-010-0014-7

Amayri O, Bouguila N (2010) A Study of Spam Filtering using Support Vector Machines. Artificial Intelligence Review 34(1):73–108, DOI 10.1007/s10462-010-9166-x, URL http://link.springer.com/10.1007/s10462-010-9166-x

Andreolini M, Bulgarelli A, Colajanni M, Mazzoni F (2005) HoneySpam : Honeypots Fighting Spam at the Source. In: Proceedings of the Steps to Reducing Unwanted Traffic on the Internet Workshop, Cambridge, MA, pp 77–83

Androutsopoulos I, Koutsias J, Chandrinos KV, Paliouras G, Spyropoulos CD (2000a) An Evaluation of Naive Bayesian Anti-Spam Filtering. In: Proceedings of 11th European Conference on Machine Learning (ECML 2000), Barcelona, pp 9–17

Androutsopoulos I, Koutsias J, Chandrinos KV, Paliouras G, Spyropoulos CD (2000b) Learning to Filter Spam E-Mail : A Comparison of a Naive Bayesian and a Memory based Approach. In: Proceedings of 4th European Conference on Principles and Practice of Knowledge Discovery in Databases, Lyon, France, September 2000, pp 1–12

Androutsopoulos I, Koutsias J, Chandrinos KV, Spyropoulos CD (2000c) An Experimental Comparison of Naive Bayesian and Keyword-Based Anti-Spam Filtering with Personal E-mail Messages. In: SIGIR '00 Proceedings of the 23rd Annual International ACM SIGIR Conference on Research and Development in Information Retrieval, pp 160–167

Androutsopoulos I, Paliouras G, Michelakis E (2006) Learning to Filter Unsolicited Commercial E-Mail. Tech. rep., National Centre for Scientific Research Demokritos, Athens, Greece

Anti-Phishing Working Group (APWG) (2014) APWG Phishing Activity Trends Report, 2nd Quarter 2014. Tech. Rep. June, Anti-Phishing Working Group

(APWG), URL http://docs.apwg.org/reports/apwg_trends_report_q2_2014.pdf

Attar A, Rad RM, Atani RE (2011) A Survey of Image Spamming and Filtering Techniques. Artificial Intelligence Review 40(1):71–105, DOI 10.1007/s10462-011-9280-4, URL http://link.springer.com/10.1007/s10462-011-9280-4

Baayen H, Halteren HV, Neijt A, Tweedie F (2002) An Experiment in Authorship Attribution. In: Proceedings of JADT 2002: Sixth International Conference on Textual Data Statistical Analysis, pp 29–37

Basavaraju M, Prabhakar R (2010) A Novel Method of Spam Mail Detection using Text Based Clustering Approach. International Journal of Computer Applications 5(4):15–25

Basnet R, Mukkamala S, Sung AH (2008) Detection of Phishing Attacks : A Machine Learning Approach. In: Studies in Fuzziness and Soft Computing, pp 373–383

Bergholz A, Beer JD, Glahn S (2010) New Filtering Approaches for Phishing Email. Journal of Computer Security 18:7–35

Beverly R, Sollins K (2008) Exploiting Transport-Level Characteristics of Spam. In: 5th Conference on Email and Anti-Spam (CEAS), Mountain View, CA

Biggio B, Fumera G, Pillai I, Roli F (2006) A Survey and Experimental Evaluation of Image Spam Filtering Techniques. Pattern Recognition Letters 32(10):1436–1446

Biggio B, Corona I, Fumera G, Giacinto G, Roli F (2011) Bagging Classifiers for Fighting Poisoning Attacks in Adversarial Classification Tasks. In: Multiple Classifier Systems, Springer Berlin Heidelberg, pp 350–359

Blanco A, Ricket AM, Martn-Merino M (2007) Combining SVM Classifiers for Email Anti-spam Filtering. In: Computational and Ambient Intelligence, Springer Berlin Heidelberg, pp 903–910

Blanzieri E, Bryl A (2008) A Survey of Learning-Based Techniques of Email Spam Filtering. Journal Artificial Intelligence Review 29(1):63–92

Boykin PO, Roychowdhury VP (2005) Leveraging Social Networks to Fight Spam. Computer 38(4):61–68

Brien CO, Vogel C (2003) Spam Filters : Bayes vs . Chi-squared ; Letters vs . Words. In: ISICT 03: Proceedings of the First International Symposium on Information and Communication Technologies, Dublin: Trinity College, September

Bringer ML, Chelmecki CA, Fujinoki H (2012) A Survey: Recent Advances and Future Trends in Honeypot Research. International Journal of Computer Network and Information Security 4(10):63–75, DOI 10.5815/ijcnis.2012.10.07, URL http://www.mecs-press.org/ijcnis/ijcnis-v4-n10/v4n10-7.html

Carpinter J, Hunt R (2006) Tightening the Net : A Review of Current and Next Generation Spam Filtering Tools. Computers and Security 25(8):566–578

Carreras X, Marquez L (2001) Boosting Trees for Anti-Spam Email Filtering p 7, URL http://arxiv.org/abs/cs/0109015, 0109015

Caruana G, Li M (2012) A Survey of Emerging Approaches to Spam Filtering. ACM Computing Surveys 44(2):1–27, DOI 10.1145/2089125.2089129, URL http://dl.acm.org/citation.cfm?doid=2089125.2089129

Chandrasekaran M, Narayanan K, Upadhyaya S (2006) Phishing E-mail Detection Based on Structural Properties. In: Proceedings of the NYS Cyber Security Conference, Albany, NY, pp 2–8

Chih-Chin Lai, Tsai MC (2004) An Empirical Performance Comparison of Machine Learning Methods for Spam E-mail Categorization. In: Fourth International Conference on Hybrid Intelligent Systems, HIS 2004, pp 0–4

Chirita PA, Diederich J, Nejdl W (2009) MailRank : Using Ranking for Spam Detection. In: Proceedings of the 14th ACM International Conference on Information and Knowledge Management, CIKM 2005, pp 373–380

Chou N, Ledesma R, Teraguchi Y, Boneh D, Mitchell JC (2004) Client-side Defense Against Web-based Identity Theft. In: Proc. 11th Annual Network and Distributed System Security Symposium (NDSS 04)

CISCO (2007) Botnets : The New Threat Landscape. Tech. rep., URL http://www.cisco.com/c/en/us/solutions/collateral/enterprise-networks/threat-control/networking_solutions_whitepaper0900aecd8072a537.pdf

CISCO (2014) Cisco 2014 Annual Security Report. Tech. rep., CISCO, URL http://www.cisco.com/web/offer/gist_ty2_asset/Cisco_2014_ASR.pdf

Clark KP (2008) A Survey of Content-based Spam Classifiers. pp 1–19

Cohen WW (1996) Learning Rules that Classify. In: Spring Symposium on Machine Learning in Information Access,, pp 18–25

Cormack GV (2008) Email Spam Filtering: A Systematic Review. Foundations and Trends in Information Retrieval 1(4):335–455, DOI 10.1561/1500000006, URL http://www.nowpublishers.com/article/Details/INR-006

Cormack GV, Lynam TR (2007) On-line Supervised Spam Filter Evaluation. ACM Transactions on Information Systems (TOIS) 25(3)

Cranor LF, Lamacchia BA (1998) Spam! Communications of the ACM 41(8)

Crocker D (2009) Internet Mail Architecture - RFC 5598. Tech. rep., URL https://tools.ietf.org/html/rfc5598

Cyberoam (2014) Internet Threats Trend Report 2014. Tech. Rep. April, Cyberoam

Debarr D, Wechsler H (2009) Spam Detection using Clustering , Random Forests , and Active Learning. In: CEAS 2009 Sixth Conference on Email and Anti-Spam

Delany SJ, Bridge D (2006) Feature based and Feature free Textual CBR : a Comparison in Spam Filtering. In: Proceedings of the 17th Irish Conference on Artificial Intelligence and Cognitive Science (AICS '06), pp 244–253

Delany SJ, Cunningham P, Tsymbal A, Coyle L (2005) A Case-based Technique for Tracking Concept Drift in Spam Filtering. Knowledge-Based Systems 18(4-5):187–195, DOI 10.1016/j.knosys.2004.10.002, URL http://linkinghub.elsevier.com/retrieve/pii/S0950705105000316

Diao Y, Lu H, Wu D (2003) A Comparative Study of Classification Based Personal E-mail Filtering. In: Knowledge Discovery and Data Mining. Current Issues and New Applications, pp 408–419

Dredze M, Schilit BN, Norvig P (2009) Suggesting Email View Filters for Triage and Search. In: Proceedings of International Joint Conference on Artificial Intelligence (IJCAI), pp 1414–1419

Drucker H, Wu D, Vapnik VN (1999) Support Vector Machines for Spam Categorization. IEEE Transactions on Neural Networks 10(5):1048–1054

Fawcett T (2004) "In vivo" Spam Filtering: A Challenge Problem for Data Mining. SIGKDD Explorations 5(2):140–148, URL http://arxiv.org/abs/cs/0405007, 0405007

Fdez-Riverola F, Iglesias E, Díaz F, Méndez J, Corchado J (2007a) Applying Lazy Learning Algorithms to Tackle Concept Drift in Spam Filtering. Expert Systems with Applications 33(1):36–48, DOI 10.1016/j.eswa.2006.04.011, URL http://linkinghub.elsevier.com/retrieve/pii/S0957417406001175

Fdez-Riverola F, Iglesias E, Díaz F, Méndez J, Corchado J (2007b) SpamHunting: An Instance-based Reasoning System for Spam Labelling and Filtering. Decision Support Systems 43(3):722–736, DOI 10.1016/j.dss.2006.11.012, URL http://linkinghub.elsevier.com/retrieve/pii/S0167923606002041

Fette I, Sadeh N, Tomasic A (2007) Learning to Detect Phishing Emails. In: Proceedings of the 16th International Conference on World Wide Web, New York, NY, USA, pp 649–656

Fumera G (2006) Spam Filtering Based On The Analysis Of Text Information Embedded Into Images. Journal of Machine Learning Research (special issue on Machine Learning in Computer Security) 7:2699–2720

Gansterer WN, Ecker GF (2008) On the Relationship Between Feature Selection and Classification Accuracy. Journal of Machine Learning Research 4:90–105

Garriss S, Kaminsky M, Freedman MJ, Karp B, Mazières D, Yu H (2006) RE : Reliable Email. In: NSDI'06 Proceedings of the 3rd Conference on Networked Systems Design & Implementation, pp 22–22

Golbeck J, Hendler J (2004) Reputation Network Analysis for Email Filtering Creating the Reputation Network. In: Proceedings of the First Conference on Email and Anti-Spam, Mountain View, California.

Gomez JC, Boiy E, Moens MF (2012) Highly Discriminative Statistical Features for Email Classification. Knowledge and Information Systems 31(1):23–53, DOI 10.1007/s10115-011-0403-7, URL http://link.springer.com/10.1007/s10115-011-0403-7

Goodman BJ, Cormack GV, Heckerman D (2007) Spam and the Ongoing Battle for the Inbox. Communications of the ACM 50(2):24–33

Graham P (2002a) A Plan for Spam. URL http://www.paulgraham.com/spam.html

Graham P (2002b) Will Filters Kill Spam? URL http://www.paulgraham.com/wfks.html

Graham P (2003) Better Bayesian Filtering. URL http://www.paulgraham.com/better.html

Graham-Cumming J (2004) How to Beat an Adaptive Spam Filter. In: The Spam Conference

Graham-Cumming J (2006) Does Bayesian Poisoning Exist? Virus Bulletin URL https://www.virusbtn.com/spambulletin/archive/2006/02/sb200602-poison.dkb?url=/archive/2006/02/sb200602-poison

Greenberg A (2010) The Most Common Words In Spam Email. URL http://www.forbes.com/sites/firewall/2010/03/17/the-most-common-words-in-spam-email/

Guerra PHC, Guedes D, Jr WM, Hoepers C, Chaves MHPC, Steding-jessen K (2010) Exploring the Spam Arms Race to Characterize Spam Evolution. In: CEAS 2010 - Seventh Collaboration, Electronic messaging, Anti-Abuse and Spam Conference, Redmond, Washington USA

Guyon I (2003) An Introduction to Variable and Feature Selection. Journal of Machine Learning Research 3:1157–1182

Guzella TS, Caminhas WM (2009) A Review of Machine Learning Approaches to Spam Filtering. Expert Systems with Applications 36(7):10,206–

10,222, DOI 10.1016/j.eswa.2009.02.037, URL http://linkinghub.elsevier.com/retrieve/pii/S095741740900181X

Gyongyi Z, Garcia-molina H (2005) Web Spam Taxonomy. In: 1st International Workshop on Adversarial Information Retrieval on the Web

Hao S, Syed NA, Feamster N, Gray AG, Krasser S (2009) Detecting Spammers with SNARE : Spatio-temporal Network-level Automatic Reputation Engine. In: Proceedings of 18th USENIX Security

Hayat MZ, Basiri J, Seyedhossein L, Shakery A (2010) Content-Based Concept Drift Detection for Email Spam Filtering. In: 5th International Symposium on Telecommunications (IST'2010), pp 531–536

He J, Thiesson B (2007) Asymmetric Gradient Boosting with Application to Spam Filtering. In: Fourth Conference on Email and Anti-Spam (CEAS), Mountain View, California, USA

Hershkop S (2006) Behavior-based Email Analysis with Application to Spam Detection. PhD thesis

Hershkop S, Stolfo SJ (2005) Identifying Spam Without Peeking at the Contents. ACM Crossroads, p 11

Hu Y, Guo C, Ngai E, Liu M, Chen S (2010) A Scalable Intelligent Non-content-based Spam-filtering Framework. Expert Systems with Applications 37(12):8557–8565, DOI 10.1016/j.eswa.2010.05.020, URL http://linkinghub.elsevier.com/retrieve/pii/S0957417410004318

IBM (2012) IBM X-Force 2012 Mid-Year Trend and Risk Report. Tech. rep., IBM, URL http://www.ibm.com/smarterplanet/global/files/ca__en_us__security__xorce_2012_midyear_trend_and_risk_report.pdf

IBM (2014) IBM X-Force Threat Intelligence Quarterly, 4Q 2014. Tech. Rep. November, IBM, URL http://public.dhe.ibm.com/common/ssi/ecm/wg/en/wgl03062usen/WGL03062USEN.PDF

Irwin B, Friedman B (2008) Spam Construction Trends. In: Information Security for South Africa (ISSA), pp 1–12

Isacenkova J, Balzarotti D (2011) Measurement and Evaluation of a Real World Deployment of a Challenge-Response Spam Filter. In: Proceedings of the 2011 ACM SIGCOMM Conference on Internet Measurement Conference (IMC '11), pp 413–426

John JP, Moshchuk A, Gribble SD, Krishnamurthy A (2009) Studying Spamming Botnets Using Botlab. In: In USENIX Symposium on Networked Systems Design and Implementation (NSDI)

Jonathan B Postel (1982) Simple Mail Transfer Protocol - RFC 281. Tech. rep., URL https://www.ietf.org/rfc/rfc0821.txt

Jorgensen Z, Zhou Y, Inge M (2008) A Multiple Instance Learning Strategy for Combating Good Word Attacks on Spam Filters. Journal of Machine Learning Research 8:1115–1146

Kanaris I, Kanaris K, Houvardas I, Stamatatos E (2006) Words vs. Character n-grams for Anti-spam Filtering. International Journal on Artificial Intelligence Tools XX(X):1–20

Kaspersky (2014) Kaspersky Security Bulletin 2014. Predictions 2015. Tech. rep.

Katakis I, Tsoumakas G, Vlahavas I (2007) Email Mining : Emerging Techniques for Email Management. In: Vakali A, Pallis G (eds) Web Data Management Practices: Emerging Techniques and Technologies, Idea Group Publishing, USA, chap 10

Kiritchenko S, Matwin S, Abu-hakima S (2004) Email Classification with Temporal Features. In: Intelligent Information Processing and Web Mining, Springer Berlin Heidelberg, pp 523–533

Kolari P, Java A, Finin T, Oates T, Joshi A (2006) Detecting Spam Blogs : A Machine Learning Approach. In: AAAI'06 Proceedings of the 21st National Conference on Artificial Intelligence, pp 1351–1356

Koprinska I, Poon J, Clark J, Chan J (2007) Learning to Classify E-mail. Information Sciences 177(10):2167–2187, DOI 10.1016/j.ins.2006.12.005, URL http://linkinghub.elsevier.com/retrieve/pii/S0020025506003707

Lai CC (2007) An Empirical Study of Three Machine Learning Methods for Spam Filtering. Knowledge-Based Systems 20(3):249–254, DOI 10.1016/j.knosys.2006.05.016, URL http://linkinghub.elsevier.com/retrieve/pii/S0950705106001390

Lee K, Caverlee J, Webb S (2010a) Uncovering Social Spammers : Social Honeypots + Machine Learning. In: Proc. of 33rd Int. ACM SIGIR Conf. on Research and Development in Information Retrieval, New York, NY, USA, pp 435–442

Lee SM, Kim DS, Kim JH, Park JS (2010b) Spam Detection Using Feature Selection and Parameters Optimization. 2010 International Conference on Complex, Intelligent and Software Intensive Systems (i):883–888, DOI 10.1109/CISIS.2010.116, URL http://ieeexplore.ieee.org/lpdocs/epic03/wrapper.htm?arnumber=5447486

Leiba B, Ossher J, Segal R, Wegman M (2005) SMTP Path Analysis. In: Proceedings of Second Conference on Email and Anti-Spam, CEAS '2005

Li F, Hsieh Mh, Gburzynski P (2007) The Community Behavior of Spammers. URL http://web.media.mit.edu/~fulu/ClusteringSpammers.pdf

Liu C, Stamm S (2007) Fighting Unicode-Obfuscated Spam

Lowd D, Meek C (2005) Good Word Attacks on Statistical Spam Filters. In: Proceedings of the Second Conference on Email and Anti-Spam (CEAS)

Ludl C, McAllister S, Kirda E, Kruegel C (2007) On the Effectiveness of Techniques to Detect Phishing Sites. In: DIMVA 07: Proceedings of the 4th International Conference on Detection of Intrusions and Malware, and Vulnerability Assessment, Springer-Verlag., Berlin, Heidelberg, pp 20–39

Ludlow M (2002) Just 150 spammers Blamed for E-mail Woe. The Sunday Times

M Sahami, S Dumais, D Heckerman, , E Horvitz (1998) A Bayesian Approach to Filtering Junk E-Mail. In: 15th National Conference on Artificial Intelligence, Madison, WI, USA, Cohen, pp 55–62

McAfee (2012) Snowshoe Spamming Emerges as Threat to Email Security. URL http://www.mcafee.com/in/security-awareness/articles/snowshoe-spamming-biggest-problem.aspx

McMillan R (2008) 100 E-mail Bouncebacks? You've Been Backscattered. URL http://www.pcworld.com/article/145449/article.html

Méndez JR, Díaz F, Iglesias EL, Corchado JM (2006) A Comparative Performance Study of Feature Selection Methods for the Anti-spam Filtering Domain. In: Advances in Data Mining. Applications in Medicine, Web Mining, Marketing, Image and Signal Mining, Springer Berlin Heidelberg, pp 106–120

Metsis V, Androutsopoulos I, Paliouras G (2006) Spam Filtering with Naive Bayes Which Naive Bayes ? In: Proceedings of the 3rd International Conference on E-mail and Anti-Spam, Mountain View, CA, USA, pp 1–5

Michelakis E, Androutsopoulos I, Paliouras G, Sakkis G (2004) Filtron : A Learning-Based Anti-Spam Filter. In: Proceedings of the 1st Conference on E-mail and Anti-Spam (CEAS)

Nattakant U (2009) Review of Browser Extensions, a Man-in-the- Browser Phishing Techniques Targeting Bank Customers. In: Proceedings of the 7th Australian Information Security Management Conference, pp 4–12

Neumayer R (2006) Clustering Based Ensemble Classification for Spam Filtering. In: Proceedings of the 6th Workshop on Data Analysis, Elfa Academic Press, pp 11–22

P Resnick (2001) Internet Message Format - RFC 2822 . Tech. Rep. April 2001, URL https://tools.ietf.org/html/rfc2822

Pantel P, Lin D (1998) SpamCop : A Spam Classification & Organization Program. In: Learning from Text Categorization Papers from the AAAI Workshop AAAI Technical Report WS-98-05, Madison, Wisconsin, pp 95–98

Perera KS, Neupane B, Faisal MA, Aung Z, Woon WL (2013) A Novel Ensemble Learning-Based Approach for Click Fraud Detection in Mobile Advertising. In: Mining Intelligence and Knowledge Exploration, Springer International Publishing, pp 370–382

Pu C, Webb S (2006) Observed Trends in Spam Construction Techniques: A Case Study of Spam Evolution. In: Proceedings of third conference on e-mail and anti-spam (CEAS), vol 6, pp 0–8

Radicati (2016) Email Statistics Report, 2012-2016 - Executive Summary. Tech. Rep. 650, Radicati

Ramachandran A, Feamster N (2006) Understanding the Network-Level Behavior of Spammers. In: Proceedings of ACM SIGCOMM

Rios G, Zha H (2004) Exploring Support Vector Machines and Random Forests for Spam Detection. In: Conference on e-mail and anti-spam (CEAS), pp 5–10

Robinson G (2003) A Statistical Approach to the Spam Problem. Linux Journal (March 2003):3, URL http://dl.acm.org/citation.cfm?id=636750.636753

Sakkis G, Androutsopoulos I, Paliouras G, Karkaletsis V (2001) Stacking Classifiers for Anti-spam Filtering of E-mail. In: Empirical methods in Natural Language Processing, pp 44–50

Sanz EP (2008) E-mail Spam Filtering. Advances in Computers 74:45–109

Sasaki M, Shinnou H (2005) Spam Detection Using Text Clustering. In: International Conference on Cyberworlds, Ml

Sculley D, Wachman GM (2007) Relaxed Online SVMs for Spam Filtering. In: SIGIR '07 Proceedings of the 30th Annual International ACM SIGIR Conference on Research and Development in Information Retrieval, pp 415–422

Seewald AK (2007) An Evaluation of Naive Bayes Variants in Content-Based Learning for Spam Filtering. Intelligent Data Analysis 11(5):497–524

Sheu Jj (2007) An Efficient Two-phase Spam Filtering Method Based on E-mails Categorization. Computers & Security 26(1):381–390

Shi L, Wang Q, Ma X, Weng M, Qiao H (2012) Spam Email Classification Using Decision Tree Ensemble. Journal of Computational Information Systems 3(February):949–956

Shi W, Xie M (2013) A Reputation-based Collaborative Approach for Spam Filtering. In: AASRI Procedia, Elsevier B.V., vol 5, pp 220–227, DOI 10.1016/j.aasri.2013.10.082, URL http://linkinghub.elsevier.com/retrieve/pii/S2212671613000838

Siefkes C, Assis F, Chhabra S, Yerazunis WS (2004) Combining Winnow and Orthogonal Sparse Bigrams

for Incremental Spam Filtering. In: European Conference on Machine Learning (ECML), pp 410–421

Siponen M, Stucke C (2006) Effective Anti-spam Strategies in Companies : An International Study. In: Proceedings of the 39th Hawaii International Conference on System Sciences (HICSS), vol 06, pp 1–10

Song Y, Kocz A, Giles CL (2009) Better Naive Bayes Classification for High-precision Spam Detection. Software Practice and Experience (April):1003–1024, DOI 10.1002/spe

Sophos (2013) Security Threat Report 2013. Tech. rep., Sophos

Sophos (2014) Security Threat Report - 2014. Tech. rep., Sophos, URL http://www.sophos.com/en-us/threat-center/medialibrary/PDFs/other/sophos-security-threat-report-2014.pdf

Soranamageswari M, Meena C (2010) Statistical Feature Extraction for Classification of Image Spam Using Artificial Neural Networks. In: 2010 Second International Conference on Machine Learning and Computing, Ieee, pp 101–105, DOI 10.1109/ICMLC.2010.72, URL http://ieeexplore.ieee.org/lpdocs/epic03/wrapper.htm?arnumber=5460761

Stepp M (2005) PhishHook : A Tool to Detect and Prevent Phishing Attacks. In: In DIMACS Workshop on Theft in E-Commerce: Content, Identity, and Service

Stern H, Mason J, Shepherd M (2004) A Linguistics-based Attack on Personalised Statistical E-mail Classiers. Tech. rep., Dalhousie Univ, URL https://www.cs.dal.ca/sites/default/files/technical_reports/CS-2004-06.pdf

Symantec (2014) Internet Security Threat Report. Tech. Rep. April, Symantec

Taylor B, Fingal D, Aberdeen D (2007) The War Against Spam : A report from the Front Line. In: Workshop on Machine Learning in Adversarial Environments for Computer Security (NIPS 2007), pp 1–3

Toolan F, Carthy J (2010) Feature Selection for Spam and Phishing Detection. In: eCrime Researchers Summit (eCrime), 2010, pp 1–12

Tretyakov K (2004) Machine Learning Techniques in Spam Filtering. In: Data Mining Problem-oriented Seminar, MTAT, May, pp 60–79

Tseng CY, Chen MS (2009) Incremental SVM Model for Spam Detection on Dynamic Email Social Networks. In: 2009 International Conference on Computational Science and Engineering, Ieee, pp 128–135, DOI 10.1109/CSE.2009.260, URL http://ieeexplore.ieee.org/lpdocs/epic03/wrapper.htm?arnumber=5284281

Tsymbal A (2004) The Problem of Concept Drift : Definitions and Related Work (Technical Report TCD-CS-2004-15). Tech. rep., Computer Science Department, Trinity College, Dublin, Ireland

Tsymbal A, Pechenizkiy M, Cunningham P (2008) Dynamic Integration of Classifiers for Handling Concept Drift Dynamic Integration of Classifiers for Handling Concept Drift. Information Fusion 9(1):56–68

Wang CC, Chen SY (2007) Using Header Session Messages to Anti-spamming. Computers & Security 26(5):381–390, DOI 10.1016/j.cose.2006.12.012, URL http://linkinghub.elsevier.com/retrieve/pii/S0167404807000065

Wang D, Irani D, Pu C (2013) A Study on Evolution of Email Spam Over Fifteen Years. In: Proceedings of the 9th IEEE International Conference on Collaborative Computing: Networking, Applications and Worksharing (CollaborateCom), Icst, Austin, TX, USA, DOI 10.4108/icst.collaboratecom.2013.254082, URL http://eudl.eu/doi/10.4108/icst.collaboratecom.2013.254082

Wang S, Wang B, Lang H, Cheng X (2005) Using Non-Textual Information to Improve Spam Filtering Performance. In: CAS-ICT at Text REtrieval Conference (TREC) 2005 SPAM Track

Whissell JS, Clarke CLA (2011) Clustering for Semi-Supervised Spam Filtering. In: Proceedings of the 8th Annual Collaboration, Electronic messaging, Anti-Abuse and Spam Conference (CEAS '11), pp 125–134

Wittel GL, Wu SF (2004) On Attacking Statistical Spam Filters. In: CEAS: First Conference on Email and Anti-Spam, Mountain View, CA

Woitaszek M, Shaaban M (2003) Identifying Junk Electronic Mail in Microsoft Outlook with a Support Vector Machine. In: Proceedings of the 2003 symposium on applications and the internet, SAINT, pp 166–169

Wu Ch (2009) Behavior-based Spam Detection using a Hybrid Method of Rule-based Techniques and Neural Networks. Expert Systems With Applications 36(3):4321–4330, DOI 10.1016/j.eswa.2008.03.002, URL http://dx.doi.org/10.1016/j.eswa.2008.03.002

Wu CH, Tsai CH (2008) Robust Classification for Spam Filtering by Back-propagation Neural Networks using Behavior-based Features. Applied Intelligence 31(2):107–121, DOI 10.1007/s10489-008-0116-0, URL http://link.springer.com/10.1007/s10489-008-0116-0

Wu X, Kumar V, Ross Quinlan J, Ghosh J, Yang Q, Motoda H, McLachlan GJ, Ng A, Liu B, Yu PS, Zhou ZH, Steinbach M, Hand DJ, Steinberg D (2007) Top 10 Algorithms in Data Mining, vol 14. DOI 10.1007/s10115-007-0114-2, URL http://link.springer.com/10.1007/s10115-007-0114-2

Xie Y, Yu F, Achan K, Panigrahy R, Hulten G, Osipkov I, Communication CC, Network N (2008) Spamming Botnets : Signatures and Characteristics. In: Proceedings of ACM SIGCOMM08, Seattle, WA

Yang J, Liu Y, Liu Z, Zhu X, Zhang X (2011) A New Feature Selection Algorithm based on Binomial Hypothesis Testing for Spam Filtering. Knowledge-Based Systems 24(6):904–914, DOI 10.1016/j.knosys.2011.04.006, URL http://linkinghub.elsevier.com/retrieve/pii/S0950705111000724

Yang J, Liu Y, Zhu X, Liu Z, Zhang X (2012) A New Feature Selection based on Comprehensive Measurement both in Inter-category and Intra-category for Text Categorization. Information Processing & Management 48(4):741–754, DOI 10.1016/j.ipm.2011.12.005, URL http://linkinghub.elsevier.com/retrieve/pii/S030645731100118X

Yang Y, Pedersen JO (1997) A Comparative Study on Feature Selection in Text Categorization. In: ICML '97 Proceedings of the Fourteenth International Conference on Machine Learning, pp 412–420

Yang Y, Yoo S, Lin F, Moon IC (2010) Personalized Email Prioritization Based on Content and Social Network Analysis. IEEE Intelligent Systems 25(4):12–18

Yeh Cy, Wu CH, Doong SH (2005) Effective Spam Classification based on Meta-Heuristics. In: Proceedings of IEEE International Conference on Systems, Man and Cybernetics, pp 3872 – 3877

Yerazunis WS (2003) Sparse Binary Polynomial Hashing and the CRM114 Discriminator Rough Guide to this Talk. In: MIT Spam Conference

Yerazunis WS (2004) The Spam-Filtering Accuracy Plateau at 99 . 9 percent Accuracy and How to Get Past It. In: MIT Spam Conference

Youn S, Mcleod D (2006) A Comparative Study for Email Classification. In: Proceedings of International Joint Conferences on Computer, Information, System Sciences, and Engineering (CISSE06), Bridgeport, CT

Yu B, Xu Zb (2008) A Comparative Study for Content-based Dynamic Spam Classification using Four Machine Learning Algorithms. Knowledge-Based Systems 21(4):355–362, DOI 10.1016/j.knosys.2008.01.001, URL http://linkinghub.elsevier.com/retrieve/pii/S0950705108000026

Zhang L, Zhu J, Yao T (2004) An Evaluation of Statistical Spam Filtering Techniques Spam Filtering as Text Categorization. ACM Transactions on Asian Language Information Processing (TALIP) 3(4):243–269

Zhu Y, Tan Y (2011) A Local-Concentration-Based Feature Extraction Approach for Spam Filtering. IEEE Trransactions on Information Forensics and Security 6(2):486–497

Zhuang L, Dunagan J, Simon DR, Wang HJ, Tygar JD (2008) Characterizing Botnets from Email Spam Records. In: LEET 08: First USENIX Workshop on Large-Scale Exploits and Emergent Threat